\pdfoutput=1

\documentclass[11pt]{article}
\usepackage[dvipsnames,svgnames]{xcolor}
\usepackage{colortbl}  

\usepackage[preprint]{acl}

\usepackage{times}
\usepackage{latexsym}

\usepackage[T1]{fontenc}

\usepackage[utf8]{inputenc}

\usepackage{microtype}

\usepackage{inconsolata}

\usepackage{graphicx}

\usepackage{booktabs}
\usepackage{multirow}
\usepackage{amsmath}

\title{Improving Arithmetic Reasoning Ability of Large Language Models through Relation Tuples, Verification and Dynamic Feedback}

\author{Zhongtao Miao, Kaiyan Zhao, Yoshimasa Tsuruoka\\
        The University of Tokyo, Tokyo, Japan \\
        \texttt{\{mzt, zhaokaiyan1006, yoshimasa-tsuruoka\}@g.ecc.u-tokyo.ac.jp}}

\begin{document}
\maketitle
\begin{abstract}
Current representations used in reasoning steps of large language models can mostly be categorized into two main types: (1) natural language, which is difficult to verify; and (2) non-natural language, usually programming code, which is difficult for people who are unfamiliar with coding to read. In this paper, we propose to use a semi-structured form to represent reasoning steps of large language models. Specifically, we use relation tuples, which are not only human-readable but also machine-friendly and easier to verify than natural language. 
We implement a framework that includes three main components: 
(1) introducing relation tuples into the reasoning steps of large language models;  
(2) implementing an automatic verification process of reasoning steps with a local code interpreter based on relation tuples; and 
(3) integrating a simple and effective dynamic feedback mechanism, which we found helpful for self-improvement of large language models. 
The experimental results on various arithmetic datasets demonstrate the effectiveness of our method in improving the arithmetic reasoning ability of large language models. The source code is available at \url{https://github.com/gpgg/art}.
\end{abstract}


\section{Introduction}

Large language models, such as GPT series~\citep{NEURIPS2020_1457c0d6,achiam2023gpt}, PaLM~\citep{anil2023palm}, Mistral~\citep{jiang2023mistral}, and LLaMA~\citep{touvron2023llama1,touvron2023llama,llama3modelcard}, have shown great success in numerous tasks that require reasoning. Besides the approach to scaling up the size of large language models and training data to enhance their reasoning ability, many prompting methods have been proposed to improve their reasoning performance.
Previous works~\citep{NEURIPS2022_9d560961,NEURIPS2022_8bb0d291,NEURIPS2022_639a9a17,pmlr-v202-gao23f}, which aim to enhance the reasoning ability of large language models, can be categorized into two main types: natural language-based approaches and non-natural language-based approaches. 
The natural language-based approaches include Chain-of-Thought (CoT)~\citep{NEURIPS2022_9d560961} and Zero-shot CoT~\citep{NEURIPS2022_8bb0d291}, which utilize intermediate reasoning steps in natural language to elicit the reasoning ability of large language models. The non-natural language-based approaches include PAL~\citep{pmlr-v202-gao23f}, which proposes to use Python code to solve math word problems.

\begin{figure}[t]
  \centering
  \includegraphics[width=0.95\columnwidth]{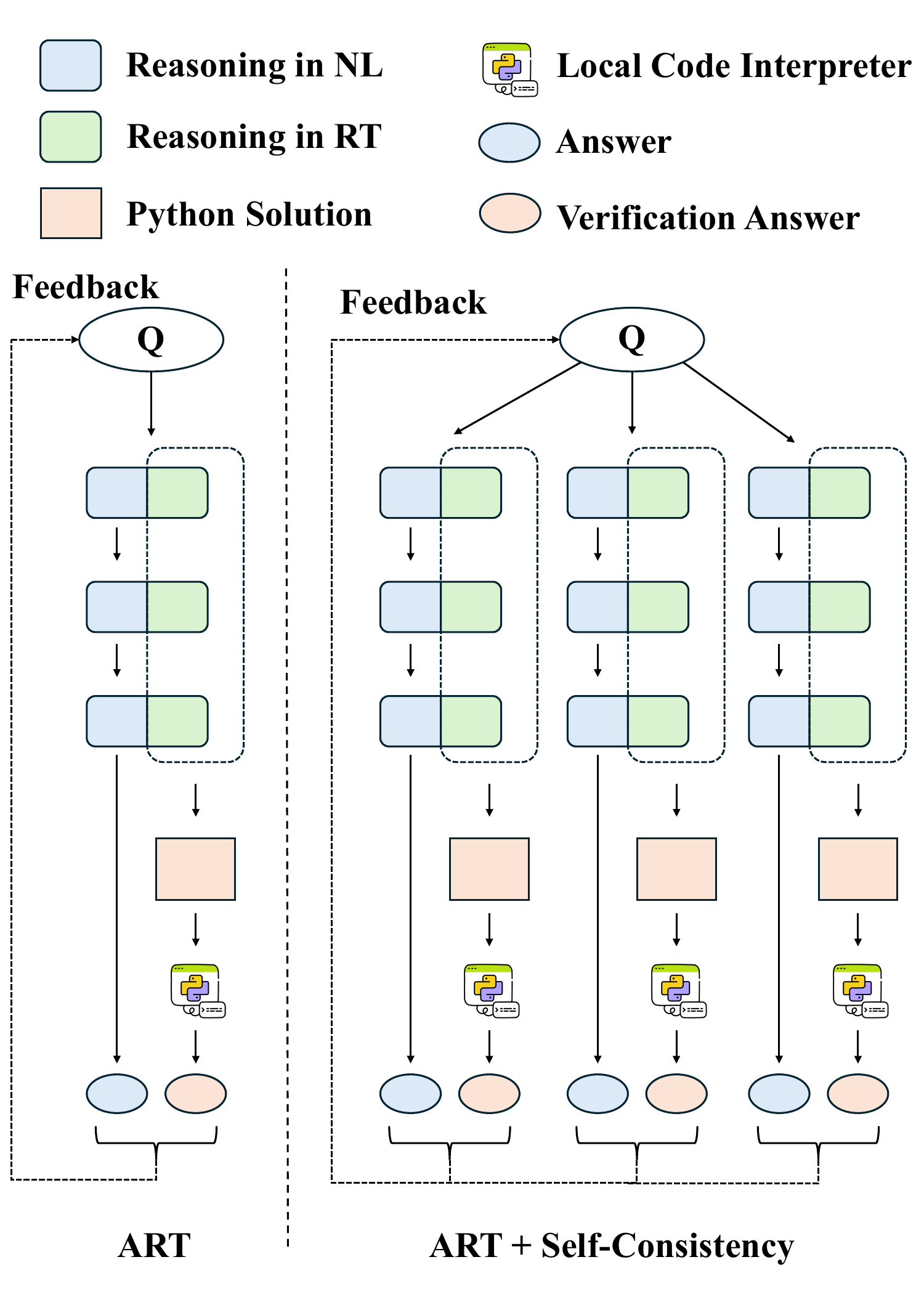}
  \caption{Schematic overview of our framework, ART.
  ``Q'' denotes a question. ``NL'' means ``Natural Language''. ``RT'' means ``Relation Tuple''. The left sub-figure shows our proposed framework ART without Self-Consistency~\citep{wang2023selfconsistency}. The right sub-figure shows that our framework can be integrated with Self-Consistency seamlessly.}
  \label{fig:ART}
\end{figure}

However, the reasoning steps represented in natural language are usually long, which can significantly increase inference cost and may contain computational errors and unjustified logical leaps~\citep{zhou2024dont}. Besides, unlike graphs or formal languages, they are difficult to verify because of the nature of natural language~\citep{zhou2024dont}.

Recently, there have been some studies that focus on translating natural language statements into formal languages such as Isabelle~\citep{10.5555/1791547} using large language models~\citep{agrawal2022towards,zhou2024dont,xu2024faithful}. However, those formal languages are hard for humans to read.

In this study, we propose a framework named ART\footnote{ART: Improving \textbf{A}rithmetic Reasoning Ability through \textbf{R}elation \textbf{T}uples, Verification and Dynamic Feedback} to enhance the arithmetic reasoning ability of large language models.
A schematic overview of our ART framework is shown in Figure~\ref{fig:ART}. First, we utilize in-context learning to make a large language model generate reasoning steps mixed with a simple semi-structured form, relation tuples. We can obtain an answer after reasoning. These relation tuples are very similar to pseudo-code, which can easily be translated into real programming code. Next, the large language model generates a Python code solution to verify the reasoning steps based on the question and relation tuples. We run the Python code in a local code interpreter to obtain the verification answer. Finally, we check whether the two answers are consistent or not and provide a dynamic feedback when necessary. If the two answers are inconsistent, we will use the large language model to regenerate a new reasoning process based on a simple dynamic feedback mechanism. The answer is determined if the two answers are consistent or reach the maximum number of tries in the feedback loop. 

\begin{figure*}[t]
  \centering
  \includegraphics[width=0.95\linewidth]{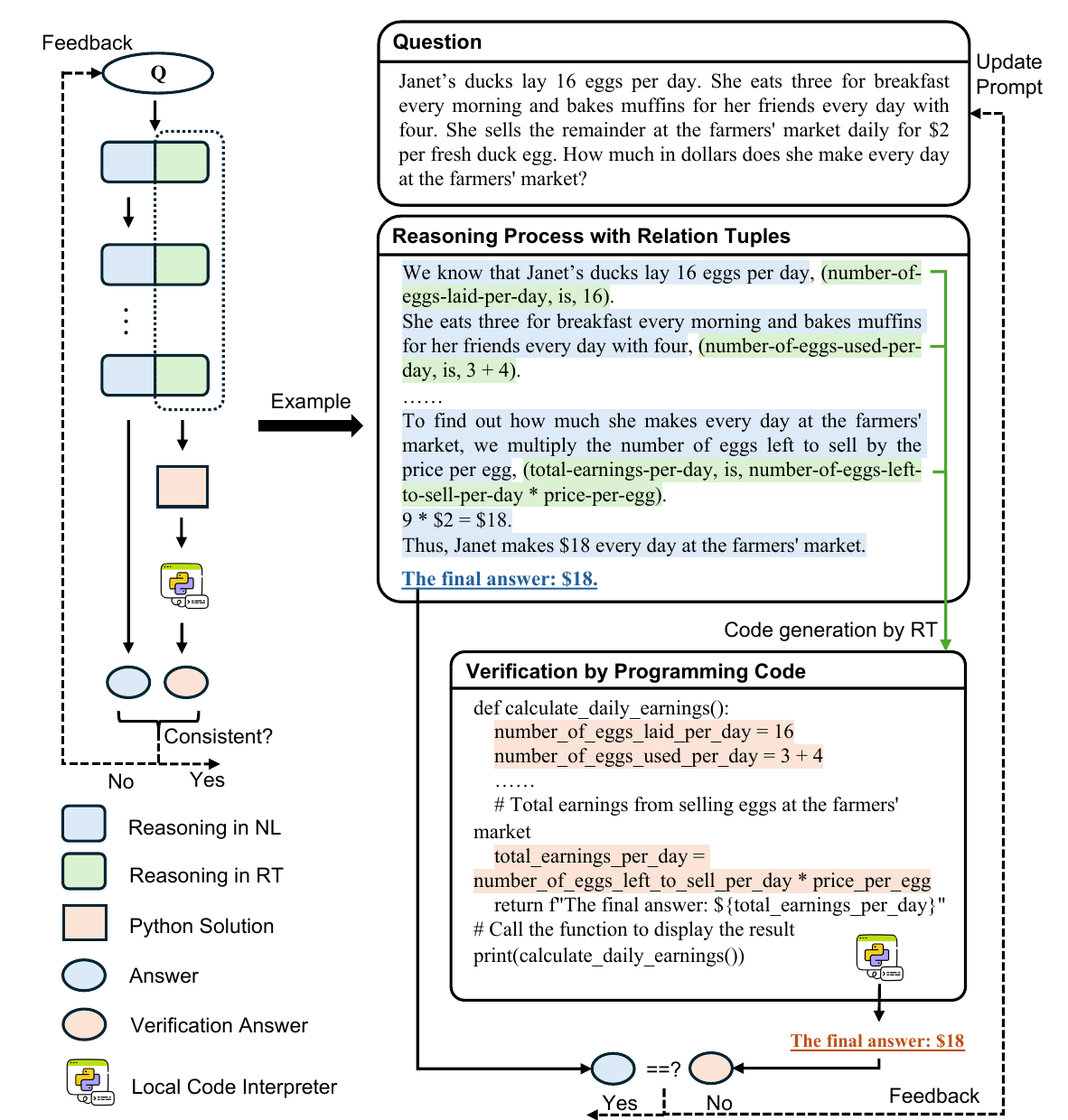}
  \caption {A detailed example illustrating how our method works. This example shows the solution to the first question of the test split of the GSM8K dataset, generated by our framework using ChatGPT.}
  \label{fig:Method}
\end{figure*}

The main contributions of this paper can be summarized as follows:
\begin{itemize}
    \item We introduce a semi-structured representation, relation tuples, into the reasoning steps of large language models. Relation tuples are usually shorter and easier to read, compared with long reasoning steps in natural language. They are more machine friendly because they are very similar to pseudo-code, which can be translated to real Python or other programming language code easily. Our findings also reveal that incorporating relation tuples into few-shot examples can improve the accuracy on four out of seven arithmetic datasets. 
    \item This study provides a local code interpreter and employs it to develop a reasoning step verifier based on relation tuples. This local code interpreter can be integrated with any large language model seamlessly, regardless of whether they are open source or not. 
    \item We implement a simple and effective dynamic feedback mechanism. Unlike Self-Refine~\citep{NEURIPS2023_91edff07}, our dynamic feedback mechanism is considerably simpler but effective. Here, ``Dynamic'' means that feedback is provided when necessary.
\end{itemize}

\section{Method}
\begin{figure}[ht]
  \centering
  \includegraphics[width=0.85\linewidth]{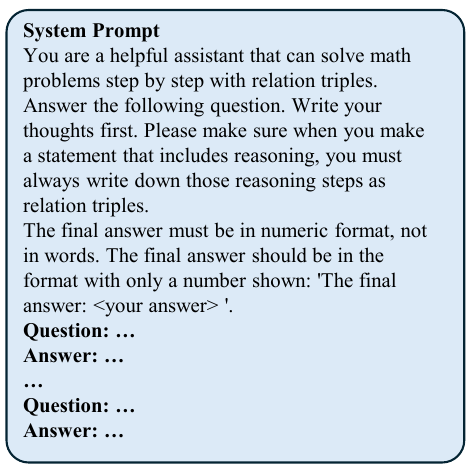}
  \caption {Prompt of relation tuple reasoning in Step 1.}
  \label{fig:RelationTuplePrompt}
\end{figure}

\subsection{Problem Formulation}
We denote a large language model as $\text{LM}$. Suppose that we have a dataset $D$. The dataset can be denoted as $D = \{Q_i, A_i\}_{i=0}^{N-1}$, where $Q_i$ is the $i$-th question, $A_i$ is the answer of $Q_i$ and $N$ is the number of examples in the dataset.
The CoT method aims to generate a series of reasoning steps and an answer, which can be denoted as:
\begin{equation}
    [\hat{R_i}, \hat{A_i}] = \text{LM}(Q_i),
\end{equation}
where $\hat{R_i}$ denotes the generated intermediate reasoning steps of the large language model $\text{LM}$ and $\hat{A_i}$ denotes the predicted answer after the reasoning steps.
The local code interpreter is denoted as $\text{LCI}$.
\subsection{ART Framework}
The ART framework can be described in the following steps:

\paragraph{Step 1: Reasoning with relation tuples.} Given a question $Q_i$ from the dataset $D$, $\text{LM}$ generates reasoning process $\hat{R_i} = \text{LM}(Q_i)$ and its answer, $\hat{A_i}$. The reasoning process consists of a series of reasoning steps and each reasoning step contains a natural language statement and its relation tuple equivalent. The reasoning process can be denoted as a list:
\begin{equation}
    \hat{R_i} = [(r_0, t_0),\dots, (r_i, t_i),\dots,(r_{n-1}, t_{n-1})],
\end{equation}
where $r_i$ is the reasoning step in natural language and $t_i$ is its equivalent in the relation tuple form. $n$ is the number of reasoning steps. The prompt used in this step is shown in Figure~\ref{fig:RelationTuplePrompt}.

\begin{figure}[t]
  \centering
  \includegraphics[width=0.85\linewidth]{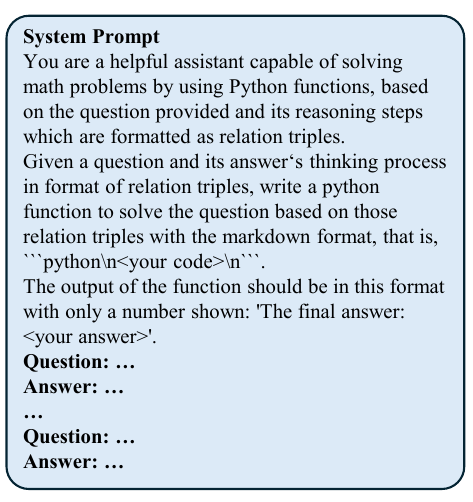}
  \caption {Prompt of program verification in Step 2.}
  \label{fig:VerificationPrompt}
\end{figure}

\paragraph{Step 2: Automatic verification with relation triples and a local code interpreter.} We can extract the relation tuples from the reasoning steps $\hat{R_i}$ in Step 1. The relation tuples extracted are denoted as a list:
\begin{equation}
    T_i = [t_0,\dots,t_i,\dots,t_{n-1}].
\end{equation}
To verify whether the reasoning steps in Step 1 are correct or not, we decide to use Python code and implement a local code interpreter. Based on the question $Q_i$ and reasoning steps in relation tuples $T_i$, $\text{LM}$ generates a Python code solution $C_i$ step by step. The code generation process can be denoted as:
\begin{equation}
    C_i = \text{LM}(Q_i, T_i).
\end{equation}
After obtaining the Python solution $C_i$. We execute it using our local code interpreter $\text{LCI}$ and get the verification answer $\hat{A_{i}^{v}}$ from the execution result:
\begin{equation}
    \hat{A_{i}^{v}} = \text{LCI}(C_i).
\end{equation}
The prompt used in this step is shown in Figure~\ref{fig:VerificationPrompt}.

\begin{figure}[t]
  \centering
  \includegraphics[width=0.85\linewidth]{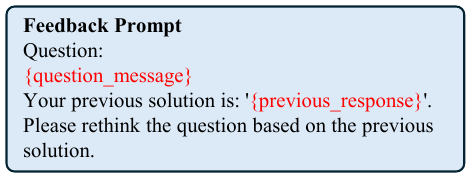}
  \caption {Feedback prompt when ART needs feedback.}
  \label{fig:FeedbackPrompt}
\end{figure}

\begin{figure*}[t]
  \centering
  \includegraphics[width=0.95\linewidth]{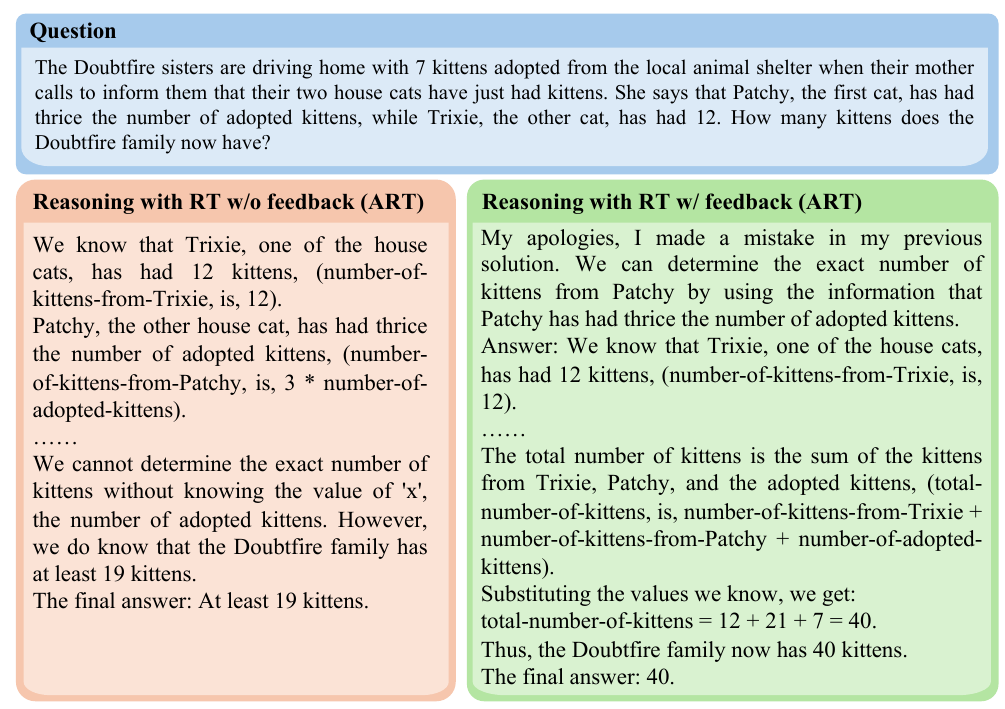}
  \caption {Comparison of ``Reasoning with RT'' solutions without feedback and with feedback, generated by our framework. This example shows the predicted solution for the $55$-th question in the test split of the GSM8K dataset, using our method with ChatGPT (\texttt{gpt-3.5-turbo-0301}).}
  \label{fig:wo_feedback_vs_w_feedback}
\end{figure*}

\paragraph{Step 3: Checking consistency and providing dynamic feedback when necessary.} From Step 1, we can get one answer $\hat{A_i}$ based on reasoning steps with relation tuples. From Step 2, we can obtain the verification answer $\hat{A_{i}^{v}}$. If these two answers are equal, it indicates that the reasoning steps in Step 1 are consistent with Step 2, confirming that there is no computational error. Therefore, the answer is determined. However, if the two answers are inconsistent, the previous reasoning steps $\hat{R_i}$ will be resent to the large language model $\text{LM}$ as a feedback. $\text{LM}$ regenerates reasoning process $\hat{R_i}$ and its answer $\hat{A_i}$ based on the feedback. The feedback prompt used here is shown in Figure~\ref{fig:FeedbackPrompt}. 
We record all the answers from Step 1 and Step 2 and choose the most common one as the final answer, ensuring seamless integration with the Self-Consistency approach~\citep{wang2023selfconsistency}.
We also provide an example to show the effectiveness of this dynamic feedback mechanism in Figure~\ref{fig:wo_feedback_vs_w_feedback}. 

\section{Experiments}

\begin{table}[t]
\centering
\resizebox{0.45\columnwidth}{!}{%
\begin{tabular}{@{}cc@{}}
\toprule
Dataset Name & \# Test Set \\ \midrule
GSM8K        & 1319        \\
ASDIV        & 2096        \\
SVAMP        & 1000        \\
SingleOP     & 562         \\
SingleEQ     & 508         \\
AddSub       & 395         \\
MultiArith   & 600         \\ \bottomrule
\end{tabular}%
}
\caption{Number of examples in the test splits of the seven arithmetic datasets we use in this study.}
\label{tab:statistics-datasets}
\end{table}

\subsection{Setup}
\paragraph{Datasets.} In this study, we focus on the arithmetic reasoning ability of large language models. We conduct experiments on seven arithmetic datasets, including GSM8K~\citep{cobbe2021gsm8k}, SVAMP~\citep{patel-etal-2021-nlp}, ASDIV~\citep{miao-etal-2020-diverse}, SingleOP, SingleEQ, AddSub and MultiArith~\citep{koncel-kedziorski-etal-2016-mawps},  following~\citet{zhao-etal-2023-automatic}.
GSM8K is a high-quality dataset which contains 8.5K problems and solutions in total. These problems usually involve 2-8 basic arithmetic operation (addition, subtraction, multiplication and division) steps to reach the final answers.
SVAMP, ASDIV, SingleOP, SingleEQ, AddSub and MultiArith are different arithmetic datasets which cover various patterns. The statistics of the datasets are shown in Table~\ref{tab:statistics-datasets}.
The primary metric used for evaluation is accuracy.





\paragraph{Models.} To evaluate our approach, we employ two proprietary large language models, ChatGPT (\texttt{gpt-3.5-turbo-0301})\footnote{\url{https://platform.openai.com/docs/deprecations/2023-06-13-updated-chat-models}} and GPT4o (\texttt{gpt-4o-2024-05-13})\footnote{\url{https://platform.openai.com/docs/models/gpt-4o}} and one open source large language model, Llama3-8B-Instruct (\texttt{meta-llama/Meta-Llama-3-8B-Instruct})\footnote{\url{https://huggingface.co/meta-llama/Meta-Llama-3-8B}}. 
Specifically, we utilize ChatGPT for our main experiments and use GPT-4o and Llama3-8B-Instruct in our ablation study.
In our framework, the temperatures of Llama3-8B-Instruct are set to 0.5 in both Step 1 and Step 2.
For the results in Table~\ref{tab:chatgpt}, both temperatures in Step 1 and Step 2 are set to 0, and the top\_p parameter is set to 1 to ensure a fair comparison with ModelSelection~\citep{zhao-etal-2023-automatic}.
We use the same prompt for all seven arithmetic datasets for each large language model.

\paragraph{In-context Learning.} For ChatGPT and Llama3-8B-Instruct, we employ in-context learning using an eight-shot setting. 
In this setting, we select the first eight questions from the train split of the GSM8K dataset.
The process of obtaining our eight-shot examples is as follows:
First, we use GPT-4 to generate CoT solutions based on the questions.
Then, we incorporate relation tuples into the reasoning steps based on the CoT solutions generated by GPT-4. 
The complete eight-shot examples are provided in Appendix~\ref{appendix:full_prompt}.
The reason for using the first eight examples of the train split of GSM8K is to avoid cherry-picking examples for in-context learning. 
For GPT-4o, following previous works~\citep{zhao-etal-2023-automatic}, we utilize a five-shot setting. The five examples are sampled from the eight-shot examples used in the eight-shot setting. Further details can be found in Appendix~\ref{appendix:full_prompt}.

\begin{table*}[ht]
\centering
\resizebox{\textwidth}{!}{%
\begin{tabular}{@{}ccccccccc@{}}
\toprule
Backbone                 & Method         & SVAMP & ASDIV & SingleOP & SingleEQ & AddSub & MultiArith & GSM8K         \\ \midrule
\multirow{4}{*}{ChatGPT} & CoT            & 83.0  & 89.3  & 94.8     & 97.4     & 90.4   & 98.7       & 80.8          \\
                         & PAL            & 80.3  & 83    & 90.7     & 97.6     & 89.4   & 96.3       & 79.2          \\
                         & ModelSelection & 84.3  & 89.4  & 94.8     & 97.8     & 90.6   & 98.7       & 82.6          \\
 & ART (ours) & \textbf{87.1} & \textbf{89.6} & \textbf{96.3} & \textbf{97.8} & \textbf{93.2} & \textbf{98.7} & \textbf{84.5} \\ \bottomrule
\end{tabular}%
}
\caption{Accuracy results on seven arithmetic datasets. The ChatGPT backbone that we use is \texttt{gpt-3.5-turbo-0301} to ensure a fair comparison with other baselines. The results of CoT, PAL and ModelSelection are quoted from~\citet{zhao-etal-2023-automatic}. Bold fonts highlight the best performance for each dataset.}
\label{tab:chatgpt}
\end{table*}

\paragraph{Implementation.} We implement our framework and conduct evaluations based on the ModelSelection codebase\footnote{\url{https://github.com/XuZhao0/Model-Selection-Reasoning}} provided by~\citet{zhao-etal-2023-automatic}.
For our local code interpreter implementation, we developed a customized version by adapting the code from Local-Code-Interpreter\footnote{\url{https://github.com/MrGreyfun/Local-Code-Interpreter}}. 
For the OpenAI Python library, we use version \texttt{1.23.2}. 
For the open source Llama3-8B-Instruct, we employ the large language model inference library \texttt{vLLM} (version 0.4.3)\footnote{\url{https://github.com/vllm-project/vllm}}~\citep{10.1145/3600006.3613165} and a single NVIDIA A100 80GB GPU to run our experiments. 

When the answers from Step 1 and Step 2 are inconsistent, the maximum number of attempts allowed in Step 3 of our framework is set to 3.

\subsection{Main Results}
As shown in Table~\ref{tab:chatgpt}, we report the accuracy results on the seven arithmetic datasets. 
Table~\ref{tab:chatgpt} shows that our approach outperforms CoT, PAL and ModelSelection baselines on ChatGPT (\texttt{gpt-3.5-turbo-0301}).
Notably, our method is particularly effective on the GSM8K, SVAMP and AddSub datasets. Specifically, it improves accuracy on the SVAMP dataset by 2.8\%, compared with ModelSelection and achieves a 1.9\% improvement over ModelSelection's 82.6\% accuracy on the GSM8K dataset.

\section{Analysis and Discussion}
\label{sec:discussion}
In this section, we analyze various factors affecting the performance of our framework. The dataset we use here is GSM8K. 
First, we investigate the effects of prior prompts used in ModelSelection~\citep{zhao-etal-2023-automatic} and GPT-4 generated prompts using the same CoT method because our eight-shot examples are created based on the GPT-4 generated solutions. Then, we assess the contributions of relation tuples, verification by programming code and feedback individually using three different large language models. Finally, we show that our method can be integrated into Self-Consistency. 


\subsection{Original Prompt vs. GPT-4 generated Prompt}
We utilize in-context learning to build our framework. Existing works use the eight-shot examples from CoT while the eight-shot examples in our method are manually created with the help of GPT-4.
Therefore, in this section, we aim to test the impact of difficulty of different prompts on the model's performance with CoT.
As shown in Table~\ref{tab:prompt_diff}, we find that the performance difference between using the two versions of prompts is not significant on ChatGPT and Llama3-8B-Instruct.
The GPT-4 generated eight-shot prompt and the eight-shot prompt used in our framework are shown in Appendix~\ref{appendix:full_prompt}.
\begin{table}[]
\centering
\resizebox{0.5\textwidth}{!}{%
\begin{tabular}{@{}ccc@{}}
\toprule
Backbone                            & Method                      & GSM8K \\ \midrule
\multirow{2}{*}{ChatGPT}            & CoT (original prompt)       & 80.8  \\
                                    & CoT (GPT-4-generated prompt) & 80.1  \\ \midrule
\multirow{2}{*}{Llama3-8B-Instruct} & CoT (original prompt)       & 80.1  \\
                                    & CoT (GPT-4-generated prompt) & 80.1  \\ \bottomrule
\end{tabular}%
}
\caption{Accuracy results on GSM8K with different eight-shot examples. The ``CoT (original prompt)'' result with ChatGPT is quoted from ModelSelection~\citep{zhao-etal-2023-automatic}}
\label{tab:prompt_diff}
\end{table}

\subsection{Role of Relation Tuples in Step 1}
In this section, we analyze the role of relation tuples. From Table~\ref{tab:cot_vs_relation_tuple}, we can observe that the reasoning process incorporating  relation tuples outperforms the CoT reasoning process on four out of the seven arithmetic datasets. 
Relation tuples in the reasoning process can be viewed as notes that record key points in the reasoning steps in natural language. These relation tuples may function as ``pause'' tokens~\citep{goyal2024think}, prompting large language models to ``think'' before generating the next reasoning step.

\begin{table*}[]
\centering
\resizebox{0.95\textwidth}{!}{%
\begin{tabular}{ccccccccc}
\hline
Backbone                 & Method               & SVAMP         & ASDIV         & SingleOP      & SingleEQ      & AddSub        & MultiArith    & GSM8K         \\ \hline
\multirow{2}{*}{ChatGPT} & CoT                  & 83.0          & \textbf{89.3} & 94.8          & \textbf{97.4} & 90.4          & \textbf{98.7} & 80.8          \\
                         & Reasoning with RT & \textbf{85.4} & 89.1          & \textbf{96.3} & 97.0          & \textbf{93.0} & 98.2          & \textbf{81.9} \\ \hline
\end{tabular}%
}
\caption{Comparison of accuracy on the seven arithmetic datasets between using prior eight-shot prompt (CoT eight-shot prompt) and using our eight-shot prompt (reasoning with RT eight-shot prompt).}
\label{tab:cot_vs_relation_tuple}
\end{table*}

\begin{table}[]
\centering
\resizebox{0.5\textwidth}{!}{%
\begin{tabular}{@{}clc@{}}
\toprule
Model                               & \multicolumn{1}{c}{Method}                 & GSM8K \\ \midrule
\multirow{4}{*}{ChatGPT}            & Reasoning with RT                          & 81.9  \\
                                    & Verification by Programming Code                     & 79.9  \\
                                    & Reasoning with RT + Verification w/o Feedback & 75.2  \\
                                    & ART (ours)                                 & 84.5  \\ \midrule
\multirow{4}{*}{Llama3-8B-Instruct} & Reasoning with RT                          & 79.6  \\
                                    & Verification by Programming Code                     & 71.6  \\
                                    & Reasoning with RT + Verification w/o Feedback & 69.1  \\
                                    & ART (ours)                                 & 80.4  \\ \midrule
\multirow{4}{*}{GPT-4o}             & Reasoning with RT                          & 96.4  \\
                                    & Verification by Programming Code                     & 95.5  \\
                                    & Reasoning with RT + Verification w/o Feedback & 95.2  \\
                                    & ART (ours)                                 & 96.6  \\ \bottomrule
\end{tabular}%
}
\caption{Accuracy results of the ablation study of our framework on the GSM8K dataset. ``RT'' means Relation Tuples. }
\label{tab:ablation_study}
\end{table}

\subsection{Role of Verification by Programming Code in Step 2}

Table~\ref{tab:ablation_study} shows the accuracy on the GSM8K dataset when using the answers from different steps of our framework as the final answers. 
In the table, ``Reasoning with RT'' represents the accuracy obtained by using the answer from Step 1 of our framework as the final answer. ``Verification by Programming Code'' indicates the accuracy achieved by using the answer from Step 2 of our framework as the final answer. The third row ``Reasoning with RT $+$ Verification w$/$o Feedback'' shows the accuracy when the two answers from Step 1 and Step 2 of our framework are consistent and correct on the first attempt.

From Table~\ref{tab:ablation_study}, it is evident that the accuracy scores on the GSM8K dataset using the verification answers from Step 2 of our framework are are lower than those using relation tuples. 
We can observe that the most obvious one is Llama3-8B-Instruct, which cannot generate programming code very well based on the semi-structured form of reasoning (relation tuples), whereas ChatGPT and GPT-4o excel in this task.

A possible reason for this discrepancy could be that in Step 2 of our framework, we use relation tuples and questions as inputs for large language models, which are infrequently encountered during their training phases. Consequently, these models struggle with generating Python solutions from this semi-structured form. 
This is particularly evident in the Llama3-8B-Instruct model, where there is an accuracy gap between using answers from Step 1 and Step 2 as final answers. This indicates that Llama3-8B-Instruct may have difficulty generating Python verification solutions based on relation tuples.

We also observe several common execution errors when Llama3-8B-instruct generates and executes Python solution code to verify the reasoning process in Step 2. Empirically, the most frequent error is ``UnboundLocalError: local variable referenced before assignment'', typically caused by using symbols that cannot serve as variable names in Python. Additionally, ``SyntaxError'' is another commonly encountered error.  
\subsection{Role of Feedback in Step 3}

\begin{figure}[t]
  \centering
  \includegraphics[width=0.9\linewidth]{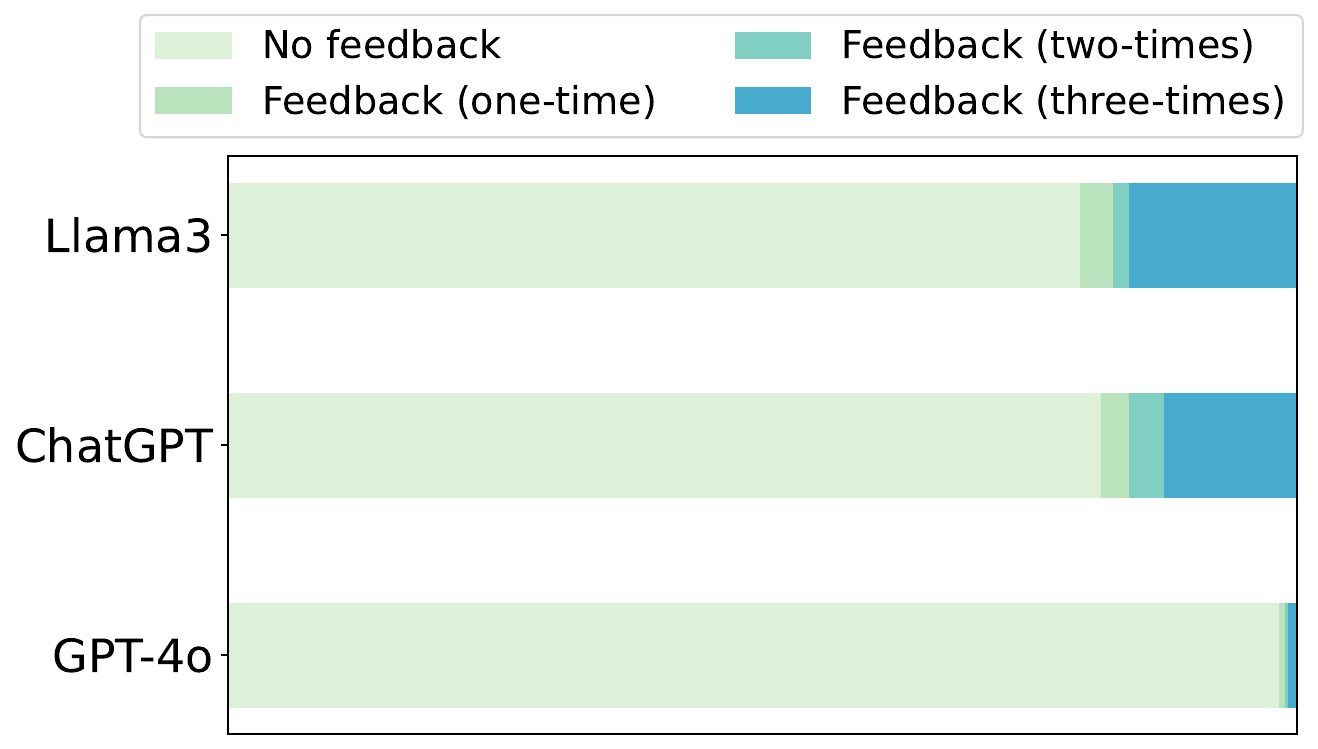}
  \caption {Percentage of questions requiring feedback on the test split of the GSM8K dataset. Note that ``Llama3'' denotes Llama3-8B-Instruct model. 
  The details can be found in Table~\ref{tab:num_questions_feedback_gsm8k}, Appendix~\ref{sec:number_of_questions_using_feedback}.}
  \label{fig:feedback_gsm8k}
\end{figure}

\begin{figure}[t]
  \includegraphics[width=\linewidth]{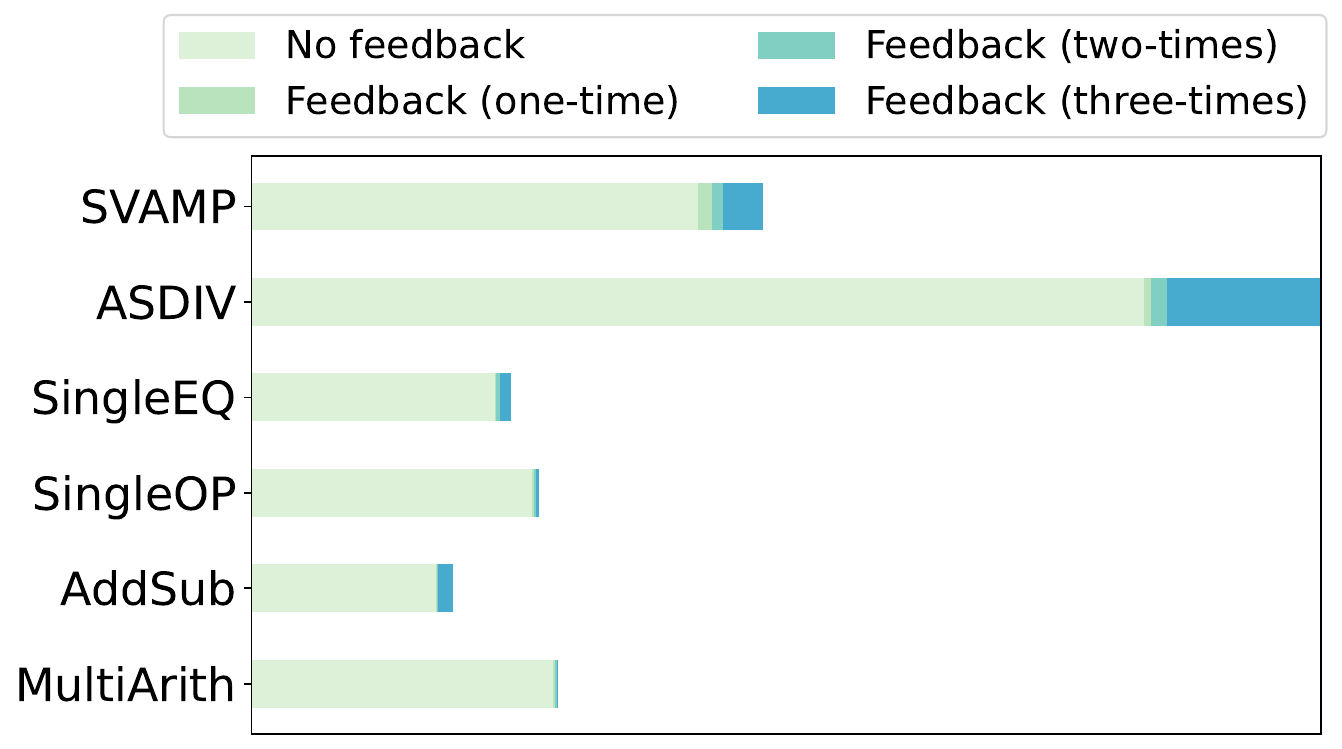}
  \caption {Percentage of questions requiring feedback on the test splits of the other six datasets (SVAMP, ASDIV, SingleEQ, SingleOP, AddSub, MultiArith) using ChatGPT (gpt-3.5-turbo-0301). The details can be found in Table~\ref{tab:num_questions_feedback_other_datasets}, Appendix~\ref{sec:number_of_questions_using_feedback}.}
  \label{fig:percentage_feedback_other_datasets}
\end{figure}

We explore the effect of the dynamic feedback mechanism in our framework in this section. Figure~\ref{fig:feedback_gsm8k} and~\ref{fig:percentage_feedback_other_datasets} show the percentage of questions utilizing feedback on the GSM8K dataset and the other 6 arithmetic datasets, respectively. 

In Figure~\ref{fig:feedback_gsm8k}, we observe an interesting phenomenon: as the coding capabilities of the large language models increase (Llama3-8B-Instruct < ChatGPT < GPT-4o) as shown in Table~\ref{tab:ablation_study}, the percentage of questions requiring feedback continuously decreases.

From Figure~\ref{fig:percentage_feedback_other_datasets}, we observe that the dataset on which  ChatGPT requires feedback most frequently is ASDIV. 
The percentage of feedback utilization might be related to the quality of datasets and the programming code understanding and generation capabilities of large language models.



\subsection{Integration with Self-Consistency}
Our framework is designed for seamless integration with the Self-Consistency approach~\citep{wang2023selfconsistency}. The core idea of Self-Consistency is to select the most common answer derived from multiple reasoning paths. In our framework, we also determine the final answer by choosing the most frequent answer from different steps.
From Table~\ref{tab:llama3_ART_SC}, we can observe that with the aid of Self-Consistency, our framework significantly enhances the arithmetic reasoning performance of Llama3-8B-Instruct on the GSM8K dataset.
\begin{table}[]
\centering
\resizebox{0.4\textwidth}{!}{%
\begin{tabular}{@{}ccc@{}}
\toprule
Backbone                            & Method                      & GSM8K \\ \midrule
\multirow{2}{*}{Llama3-8B-Instruct} 
                                    & ART (SC@1)                  & 80.4  \\
                                    & ART (SC@5)                  & 84.2  \\ \bottomrule
\end{tabular}%
}
\caption{Accuracy on the GSM8K dataset after integrating Self-Consistency (SC@5) into our framework ART. ``SC@5'' means that the number of sampled paths is 5.}
\label{tab:llama3_ART_SC}
\end{table}

\section{Related Work}
\paragraph{Natural language reasoning.} There are large amounts of studies~\citep{qiao-etal-2023-reasoning,sanyal-etal-2022-fairr,nye2021show,wang-etal-2022-iteratively} focusing on enhancing the reasoning ability of large language models in natural language. Chain-of-Thought (CoT)~\citep{NEURIPS2022_9d560961} shows that intermediate reasoning steps can improve the performance of large language models. Zero-shot CoT, as proposed by~\citet{NEURIPS2022_8bb0d291}, involves simply adding ``Let’s think step by step'' before generating answers to elicit the reasoning ability of large language models. 
Least-to-most prompting~\citep{zhou2023leasttomost} breaks down complex problems to simpler problems and solve them in sequence to enable complex reasoning in large language models. Self-Consistency~\citep{wang2023selfconsistency} extends CoT by sampling various reasoning paths, generating multiple answers and choosing the most common one. Tree-of-Thought~\citep{NEURIPS2023_271db992} generalizes over Chain-of-Thought by framing any problem as a search over a tree. 
\citet{besta2024got} propose Graph-of-Thoughts to improve large language model's reasoning ability by modeling large language model thoughts as vertices and dependencies between these vertices as edges.
Buffer-of-Thoughts~\citep{yang2024buffer} is a novel prompting approach which employs a meta-buffer to store a series of thought templates~(high-level thoughts) and retrieves a relevant thought template and instantiate it when conducting reasoning.

\paragraph{Non-natural language reasoning and verification.}
There are many works~\citep{kadlcik-etal-2023-calc,pmlr-v202-gao23f,xu2024faithful} aiming to enhance the reasoning ability of large language models by using non-natural language forms during the reasoning process.
PAL~\citep{pmlr-v202-gao23f} employs large language models to generate Python code as intermediate reasoning steps.
ERA-CoT~\citep{liu2024era} aids large language models in reasoning by analyzing entities and relationships in natural language statements. 
\citet{zhou2024solving} find that GPT-4's powerful skills in generating and executing code could be utilized to enhance mathematical reasoning ability by analyzing the Code Usage Frequency of the GPT-4 Code Interpreter.
MathCoder~\citep{wang2024mathcoder} integrates natural language reasoning, code generation and execution results to enhance the mathematical reasoning ability of large language models by fine-tuning them. 
SymbolCoT~\citep{xu2024faithful} integrates symbolic expressions and logic rules into the reasoning process of large language models to enhance their logical reasoning ability.
\citet{zhou2024dont} translate informal natural language reasoning statements into formal Isabelle code which can be verified automatically to enhance internal consistency of reasoning in large language models. Different from these works, our method utilizes the semi-structure understanding and code generation ability of large language models to verify the reasoning process.
\paragraph{Self-improvement and verification.}
There are many works focusing on the self-improvement of large language models~\citep{huang-etal-2023-large,NEURIPS2023_91edff07,haluptzok2023language,xu2024wizardlm,yu2023teaching}.
\citet{NEURIPS2022_639a9a17} propose Self-Taught Reasoner (STaR), which employs a reasoning process generation loop to produce reasoning steps and use these generated reasoning paths whose final answers are correct to further fine-tune large language models.
\citet{NEURIPS2023_91edff07} propose Self-Refine, which has three components (generator, feedback provider and refiner). Compared with Self-Refine, the dynamic feedback in our framework is provided only when necessary. Moreover, our framework does not need the feedback provider.

\section{Conclusion}
In this paper, we propose to use a semi-structured representation for the arithmetic reasoning steps of large language models. Specifically, we utilize relation tuples to connect reasoning in natural language with formal languages, such as programming code, to more effectively verify the reasoning process of large language models. These relation tuples are human-readable and can easily be translated into formal languages.

Based on this new representation of reasoning steps, we have implemented a novel framework that integrates the semi-structured representation, relation tuples, into the reasoning process of large language models. Additionally, we developed a local code interpreter to verify the reasoning process of large language models. Our framework also includes a simple and effective dynamic feedback mechanism to elicit the self-improvement ability of large language models. Experimental results demonstrate that our framework can improve the arithmetic reasoning ability of large language models.

\section*{Limitations}
We utilize programming code based on relation tuples to verify reasoning process. Therefore, our method highly depends on the programming code understanding and generation ability of large language models that we use.

Besides, the reasoning process in our method is a mixture of informal natural language statements and semi-structured relation tuples. Therefore, the inference cost is high. It will be great if large language models can reason with relation tuples only, which can reduce the inference cost while maintaining readability and are easy for machine to further process these relation tuples (e.g., automatic verification).

Finally, there might be other semi-structured forms of reasoning steps which are easy to verify. 

\section*{Ethics Statement}
This research aims to improve arithmetic reasoning ability of large language models by introducing a semi-structured form into reasoning process of large language models, a verification process and a dynamic feedback mechanism. 
We utilized publicly available datasets compiled from other research papers. No personal data was used in this study.
We agree to the License Terms and Privacy Policy of corresponding large language models and datasets used in our study.
Our research adheres to ethical AI principles, promoting the beneficial use of AI.
In addition, large language models may generate harmful contents which we are trying to avoid.
We employ GitHub Copilot to help with coding our experiments.

\bibliography{anthology,custom}

\clearpage

\appendix
\section{Comparison of different methods}
\begin{figure*}[t]
  \centering
  \includegraphics[width=0.95\linewidth]{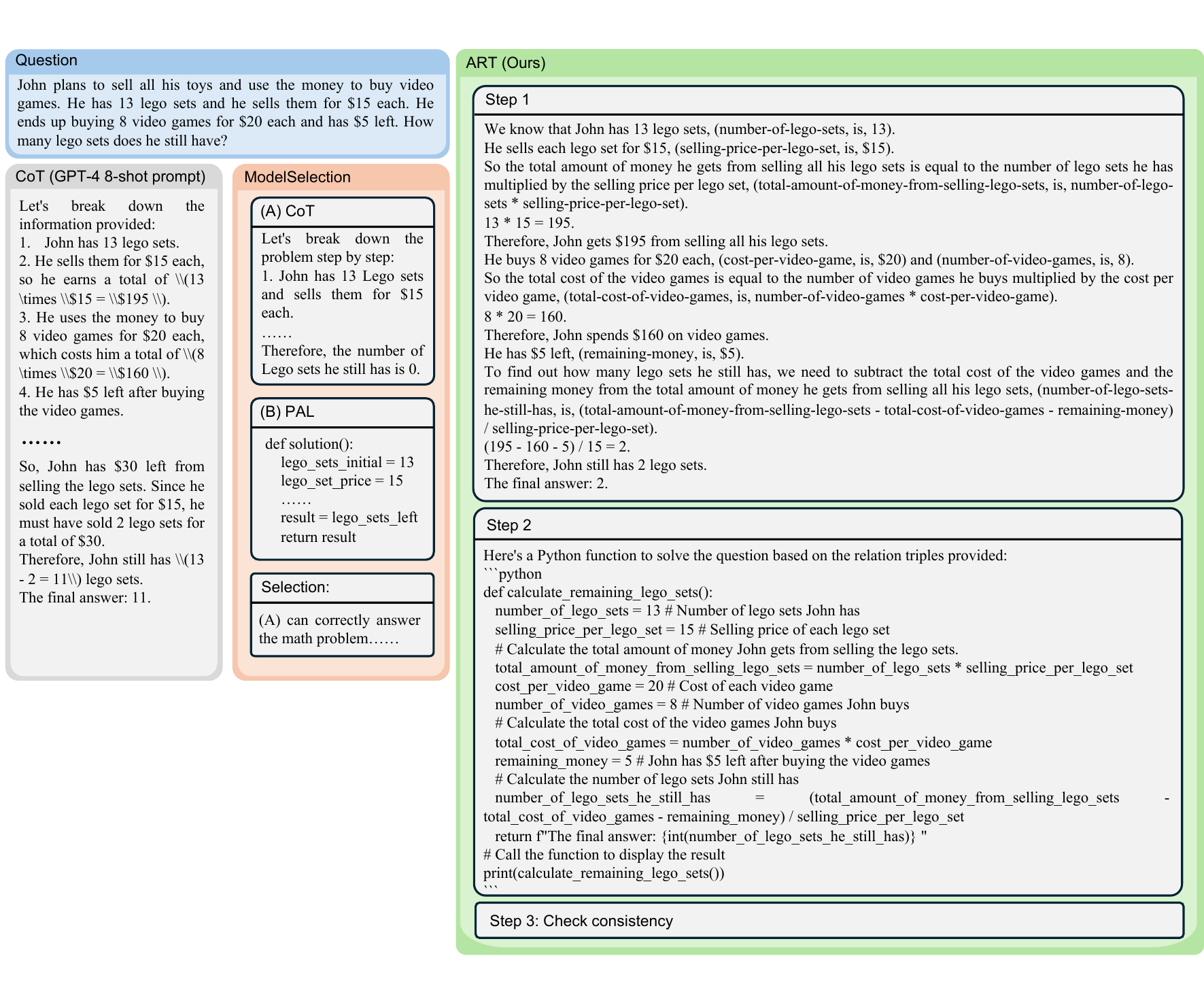}
  \caption {Comparison of different methods.}
  \label{fig:Comparison}
\end{figure*}

A comparison of different methods is shown in Figure~\ref{fig:Comparison}.

\section{Full Prompts}
\label{appendix:full_prompt}
\subsection{Eight-shot examples}
\begin{figure*}[h]
  \centering
  \includegraphics[width=1\linewidth]{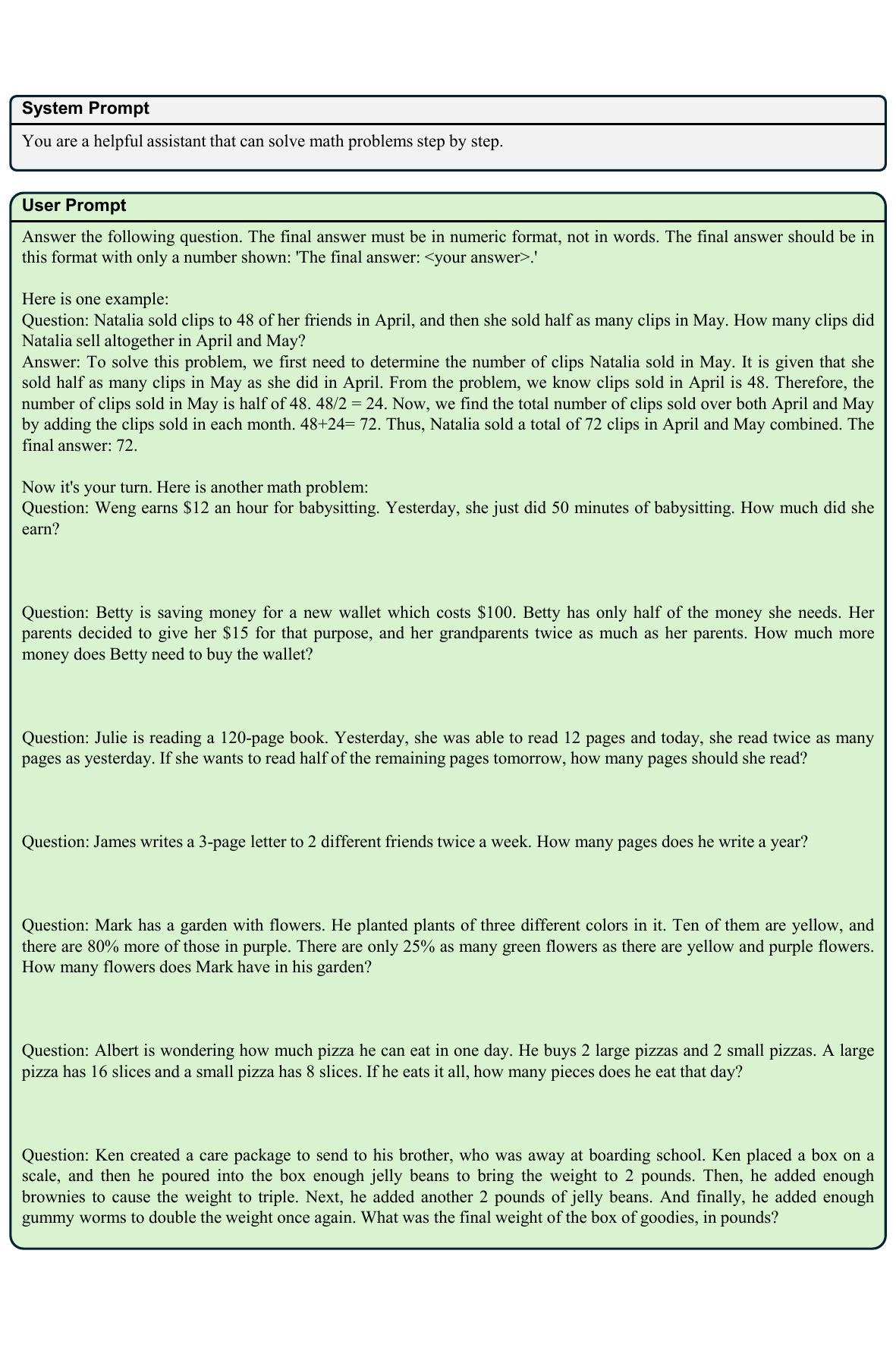}
  \caption {The system prompt and user prompt of the eight-shot example of CoT (GPT-4-Generated Prompt).}
  \label{fig:cot_gpt4_generated_01}
\end{figure*}

\begin{figure*}[h]
  \centering
  \includegraphics[width=1\linewidth]{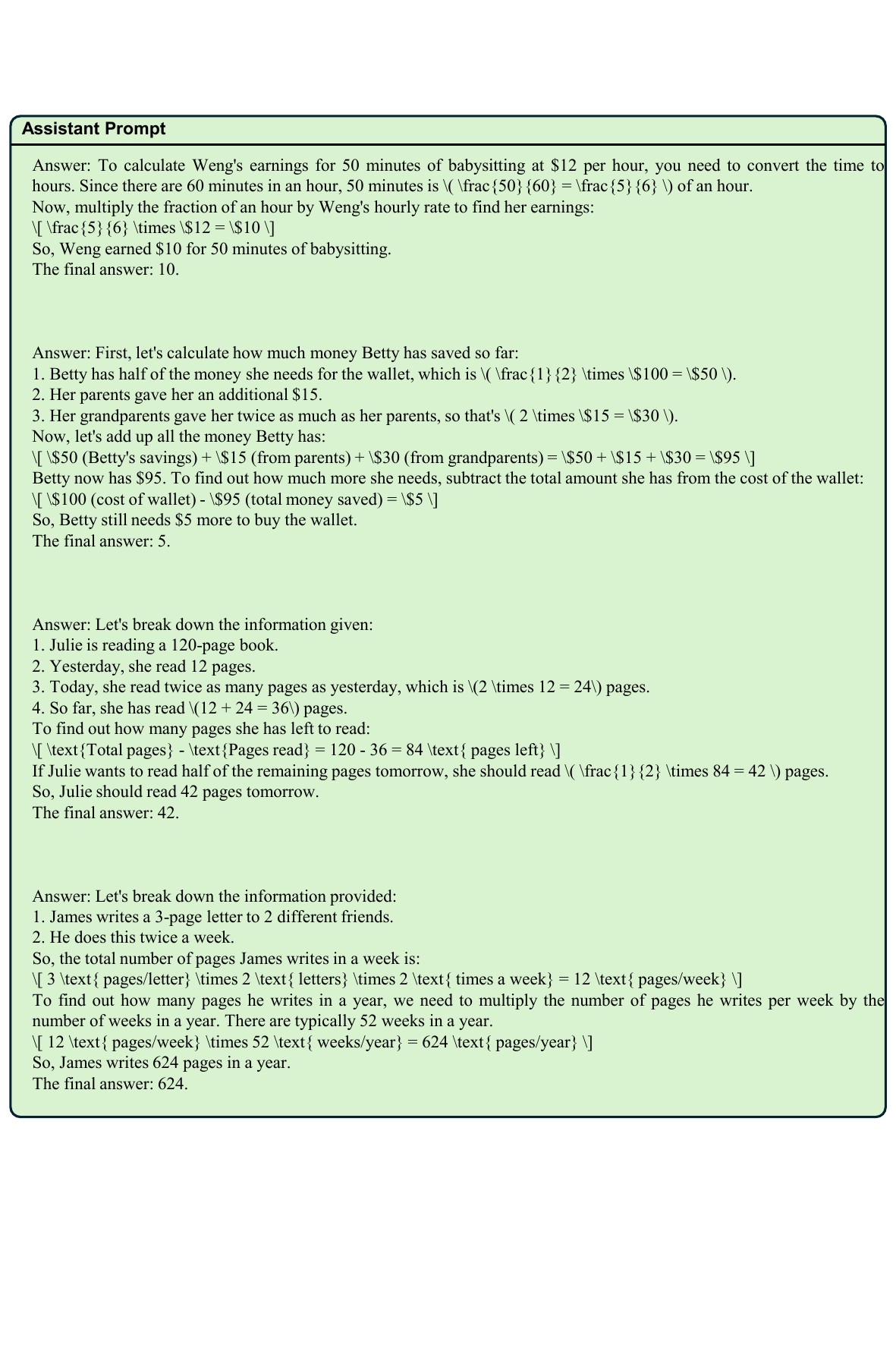}
  \caption {The assistant prompt of the eight-shot example of CoT (GPT-4-Generated Prompt).}
  \label{fig:cot_gpt4_generated_02}
\end{figure*}

\begin{figure*}[h]
  \centering
  \includegraphics[width=1\linewidth]{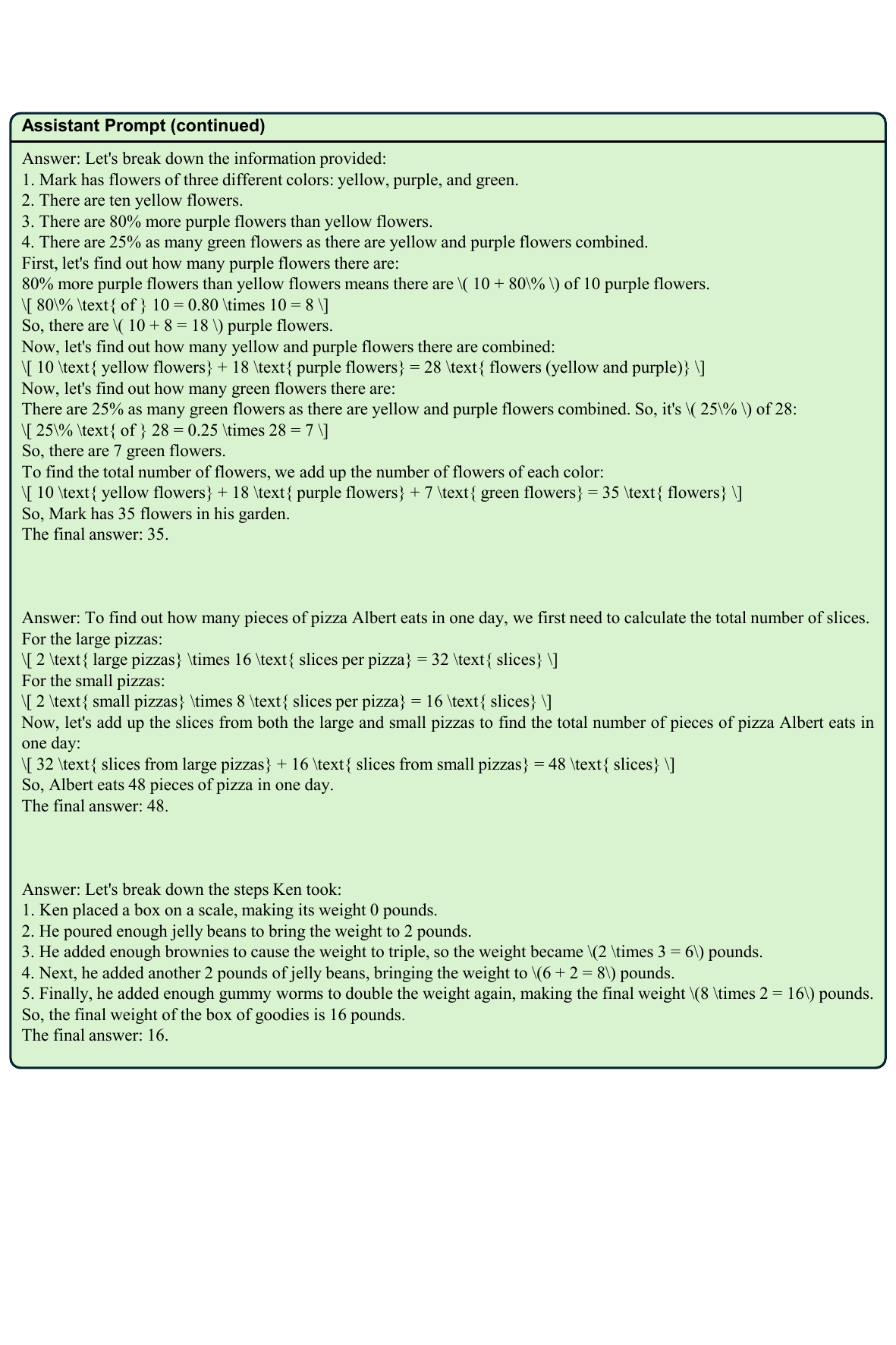}
  \caption {The assistant prompt of the eight-shot example of CoT (GPT-4-Generated Prompt) (continued).}
  \label{fig:cot_gpt4_generated_03}
\end{figure*}

Figure~\ref{fig:cot_gpt4_generated_01}, Figure~\ref{fig:cot_gpt4_generated_02} and Figure~\ref{fig:cot_gpt4_generated_03} show the full prompts of CoT (GPT-4-generated Prompt).

\begin{figure*}[h]
  \centering
  \includegraphics[width=1\linewidth]{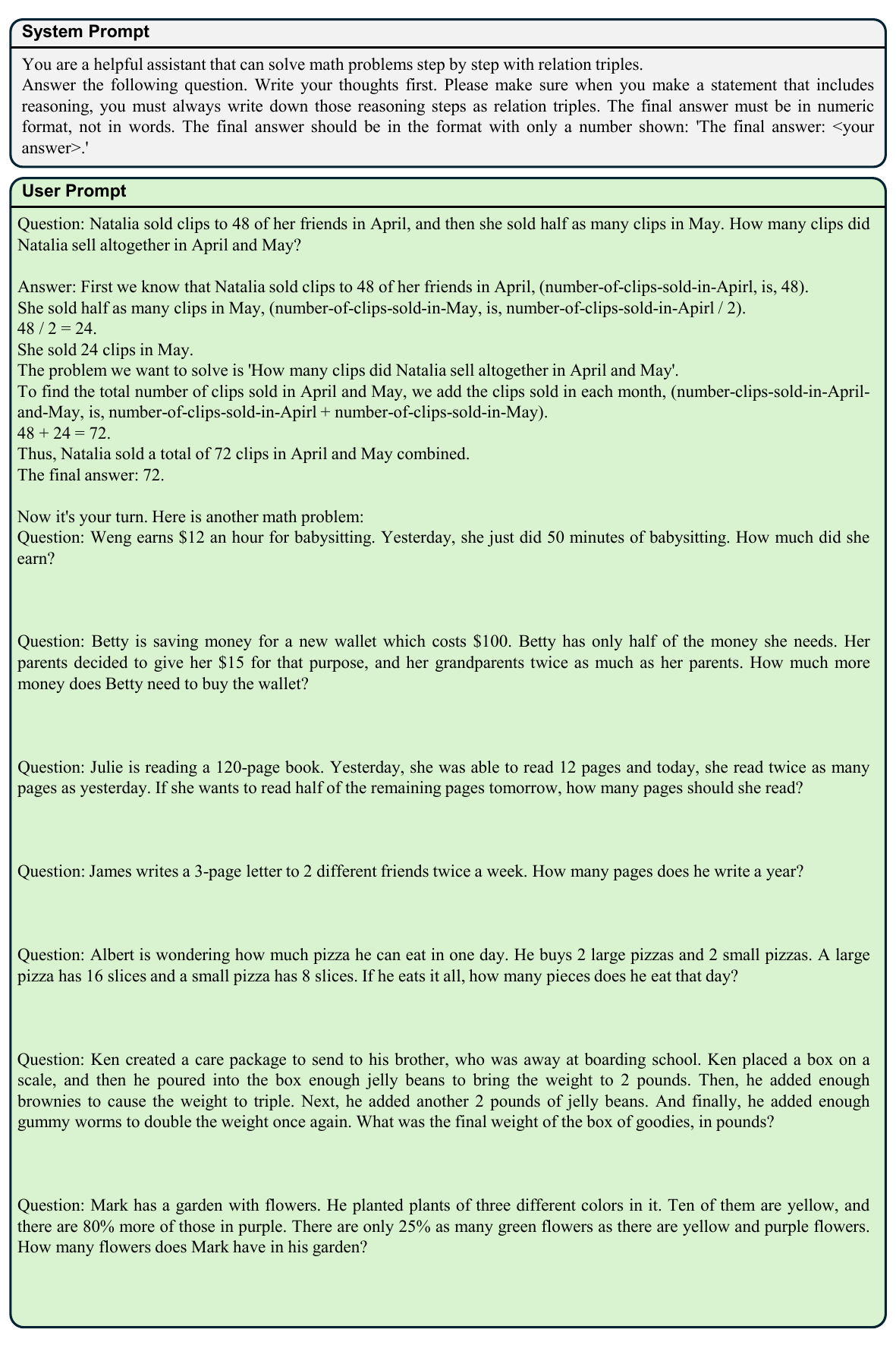}
  \caption {The system prompt and user prompt of the eight-shot example in step 1 of our framework.}
  \label{fig:step1_prompt_01}
\end{figure*}

\begin{figure*}[h]
  \centering
  \includegraphics[width=1\linewidth]{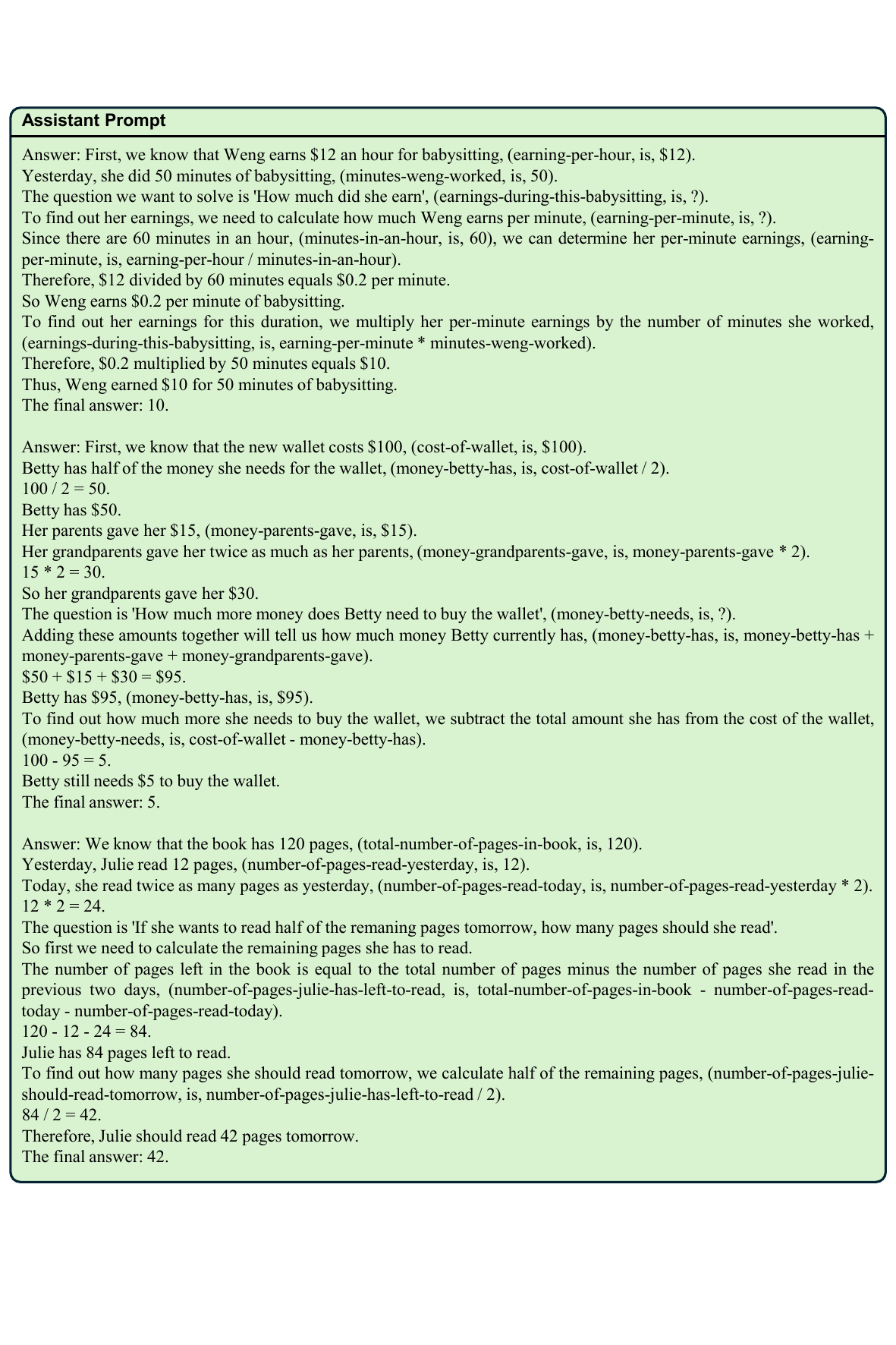}
  \caption {The assistant prompt of the eight-shot example in step 1 of our framework.}
  \label{fig:step1_prompt_02}
\end{figure*}

\begin{figure*}[h]
  \centering
  \includegraphics[width=1\linewidth]{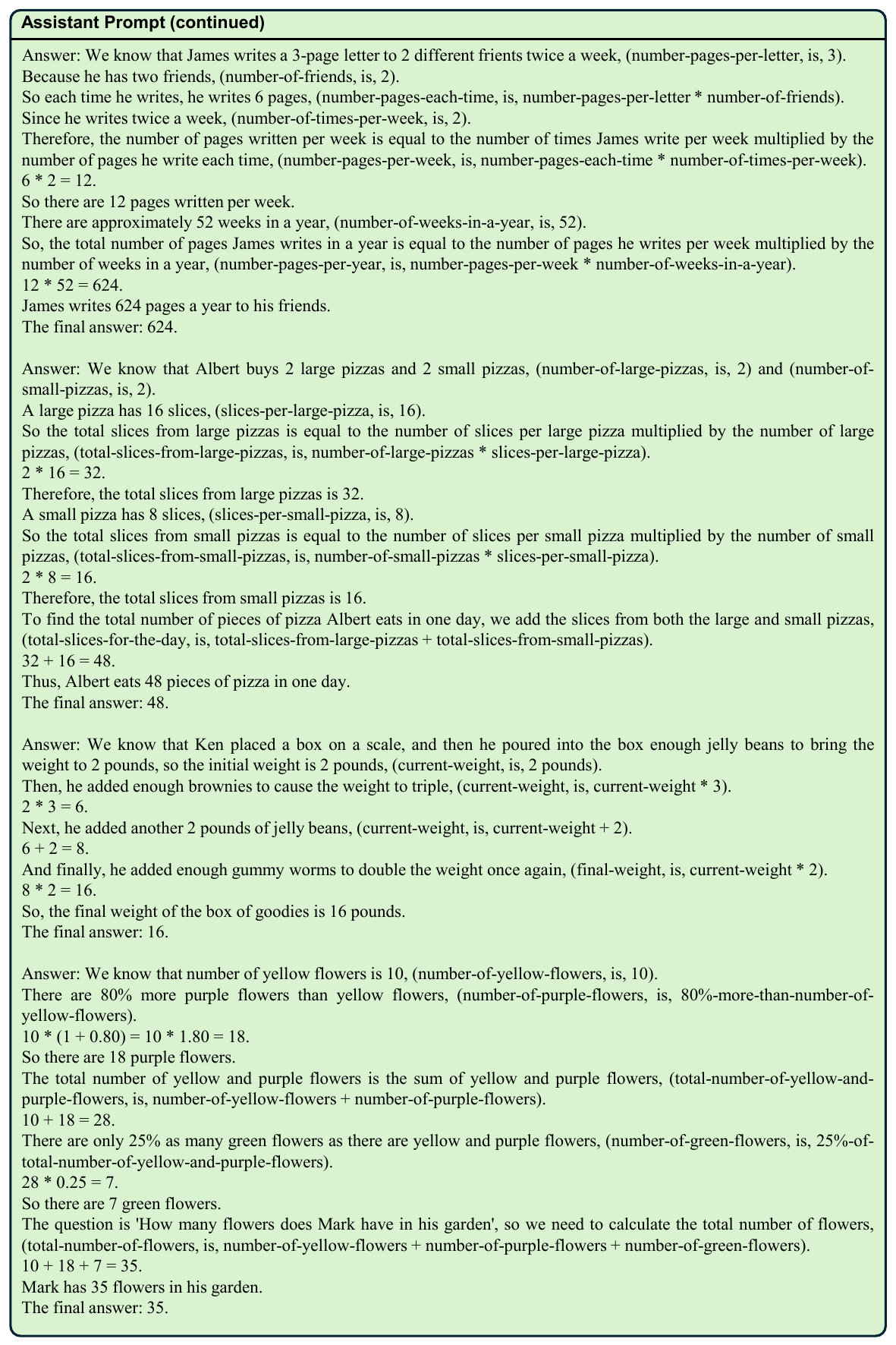}
  \caption {The assistant prompt of the eight-shot example in step 1 of our framework (continued).}
  \label{fig:step1_prompt_03}
\end{figure*}

Figure~\ref{fig:step1_prompt_01}, Figure~\ref{fig:step1_prompt_02} and Figure~\ref{fig:step1_prompt_03} show the full prompts used in step 1 of our framework in the eight-shot setting.

\begin{figure*}[h]
  \centering
  \includegraphics[width=1\linewidth]{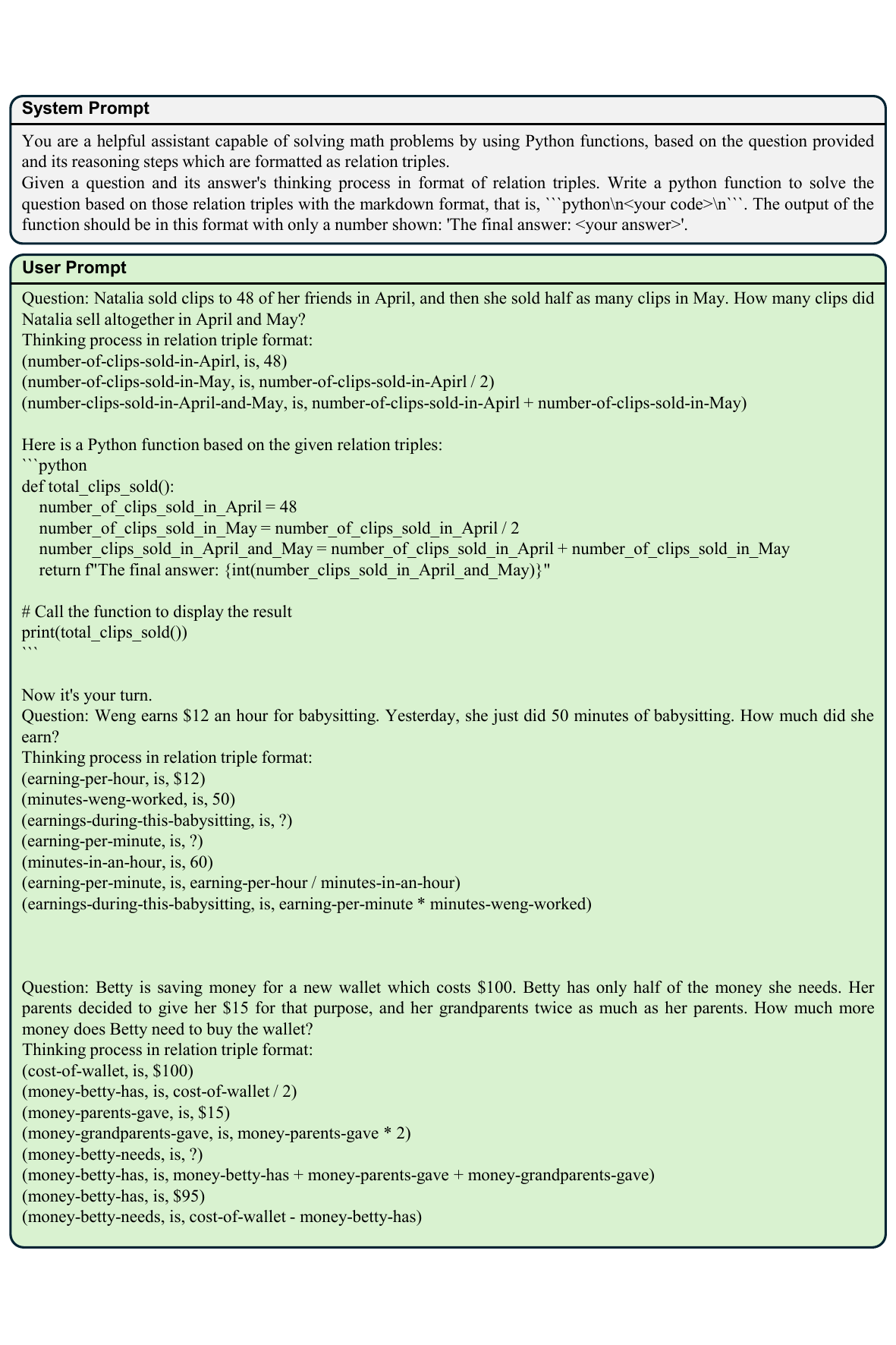}
  \caption {The system prompt and user prompt of the eight-shot example in step 2 of our framework.}
  \label{fig:step2_prompt_01}
\end{figure*}

\begin{figure*}[h]
  \centering
  \includegraphics[width=1\linewidth]{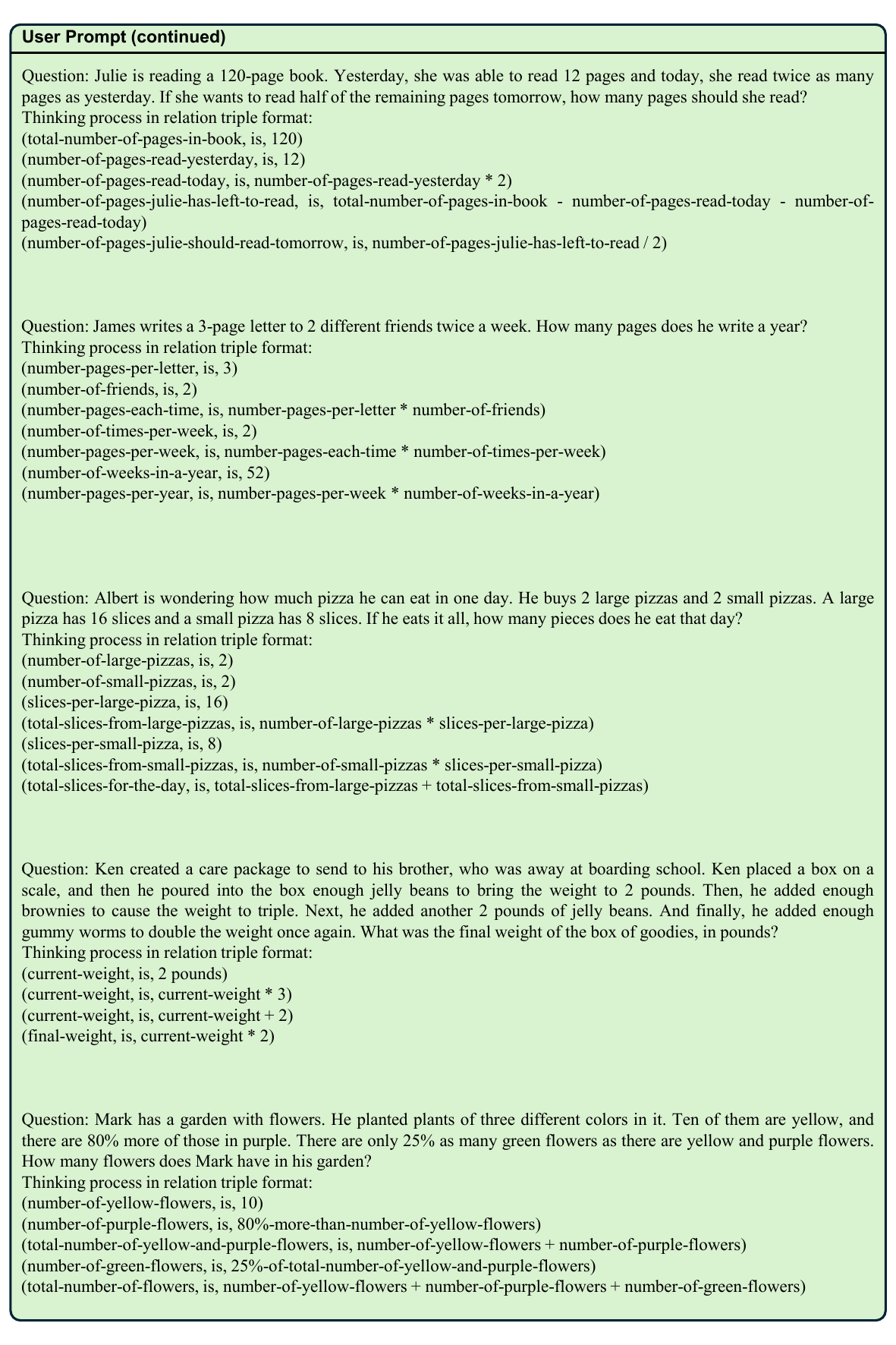}
  \caption {The user prompt of the eight-shot example in step 2 of our framework (continued).}
  \label{fig:step2_prompt_02}
\end{figure*}

\begin{figure*}[h]
  \centering
  \includegraphics[width=1\linewidth]{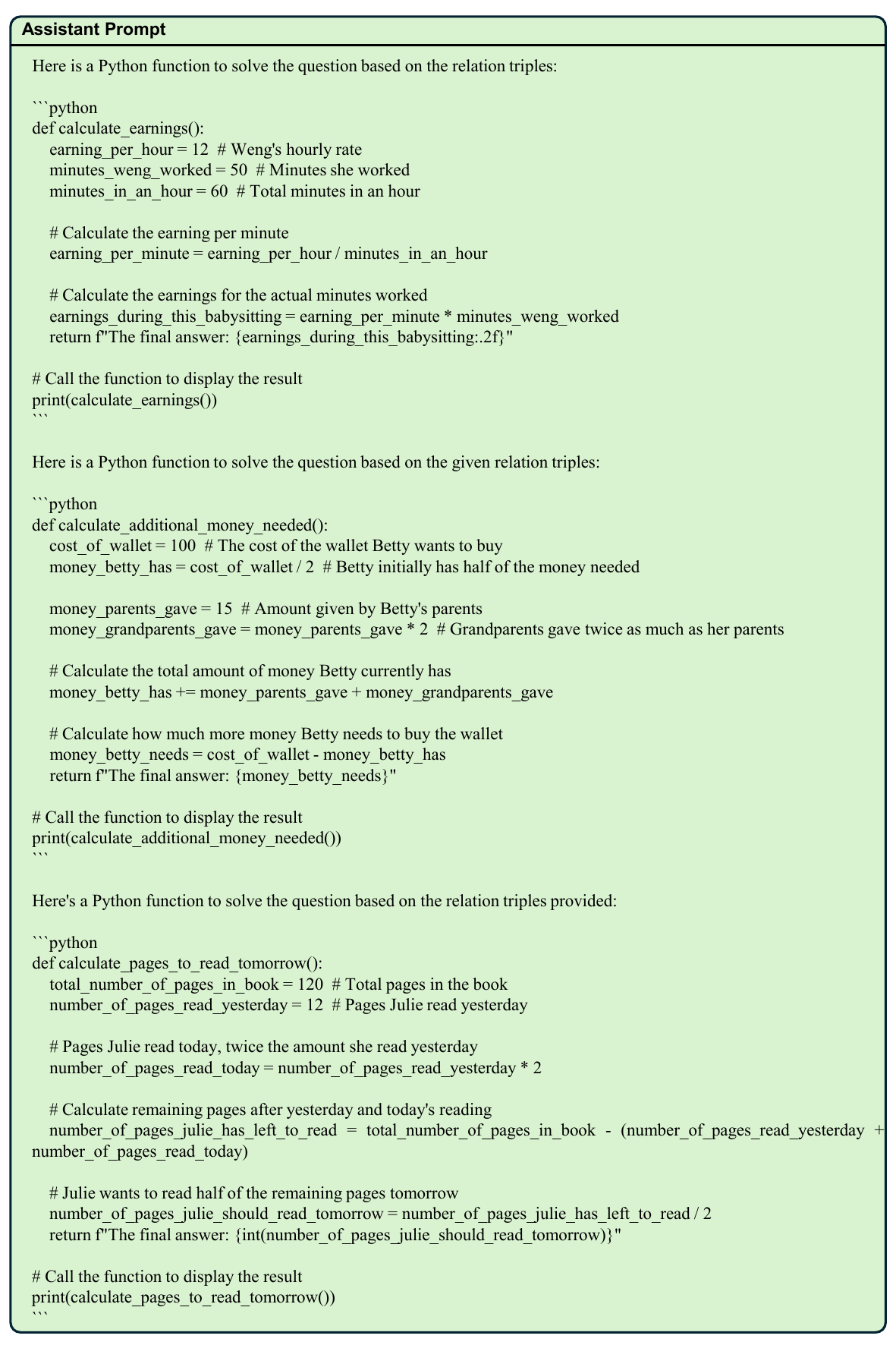}
  \caption {The assistant prompt of the eight-shot example in step 2 of our framework.}
  \label{fig:step2_prompt_03}
\end{figure*}

\begin{figure*}[h]
  \centering
  \includegraphics[width=1\linewidth]{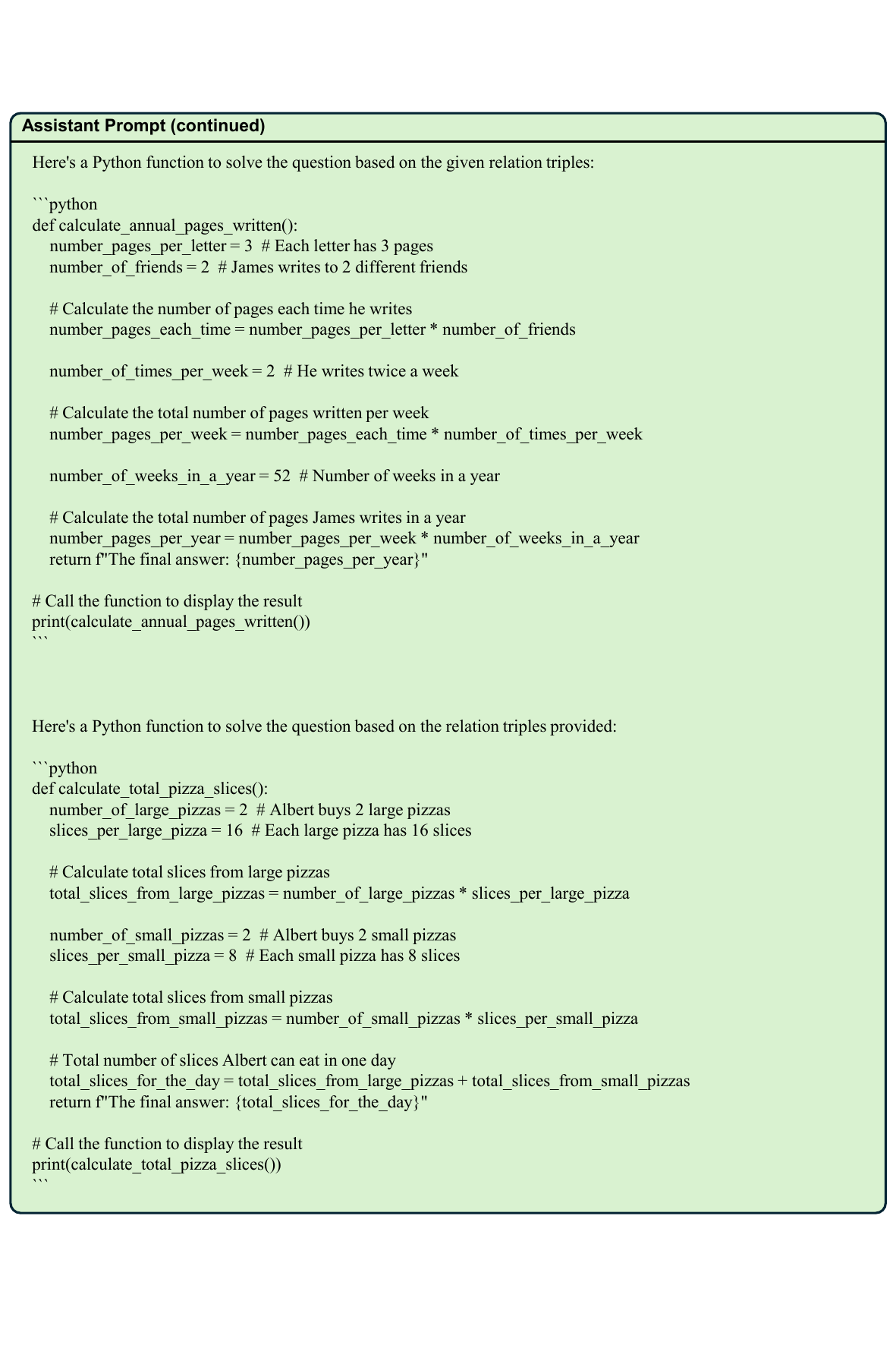}
  \caption {The assistant prompt of the eight-shot example in step 2 of our framework (continued).}
  \label{fig:step2_prompt_04}
\end{figure*}

\begin{figure*}[h]
  \centering
  \includegraphics[width=1\linewidth]{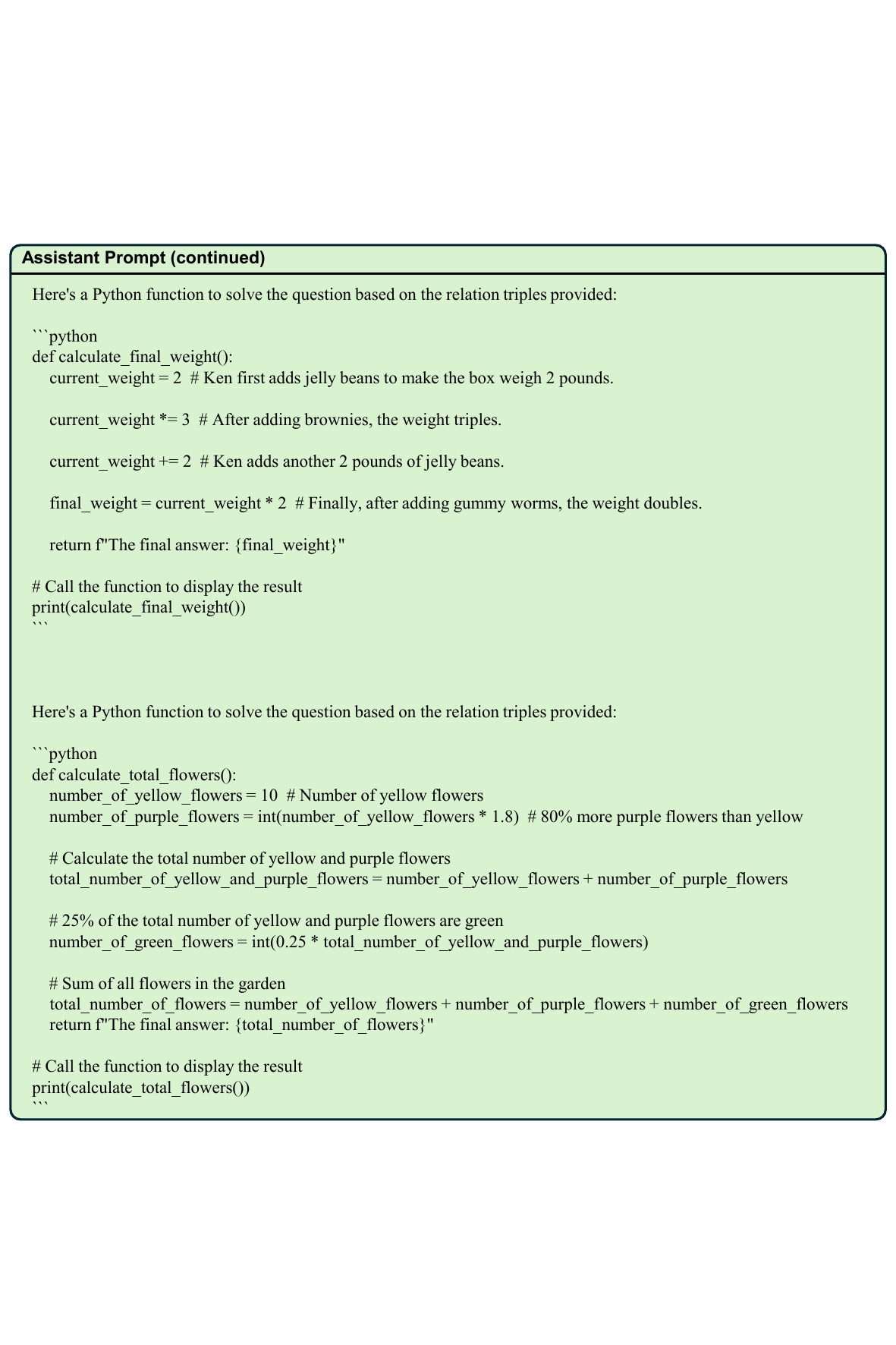}
  \caption {The assistant prompt of the eight-shot example in step 2 of our framework (continued).}
  \label{fig:step2_prompt_05}
\end{figure*}

Figure~\ref{fig:step2_prompt_01}, Figure~\ref{fig:step2_prompt_02},  Figure~\ref{fig:step2_prompt_03}, Figure~\ref{fig:step2_prompt_04} and Figure~\ref{fig:step2_prompt_05} show the full prompts used in step 2 of our framework in the eight-shot setting.

\subsection{Five-shot examples}
\begin{figure*}[h]
  \centering
  \includegraphics[width=1\linewidth]{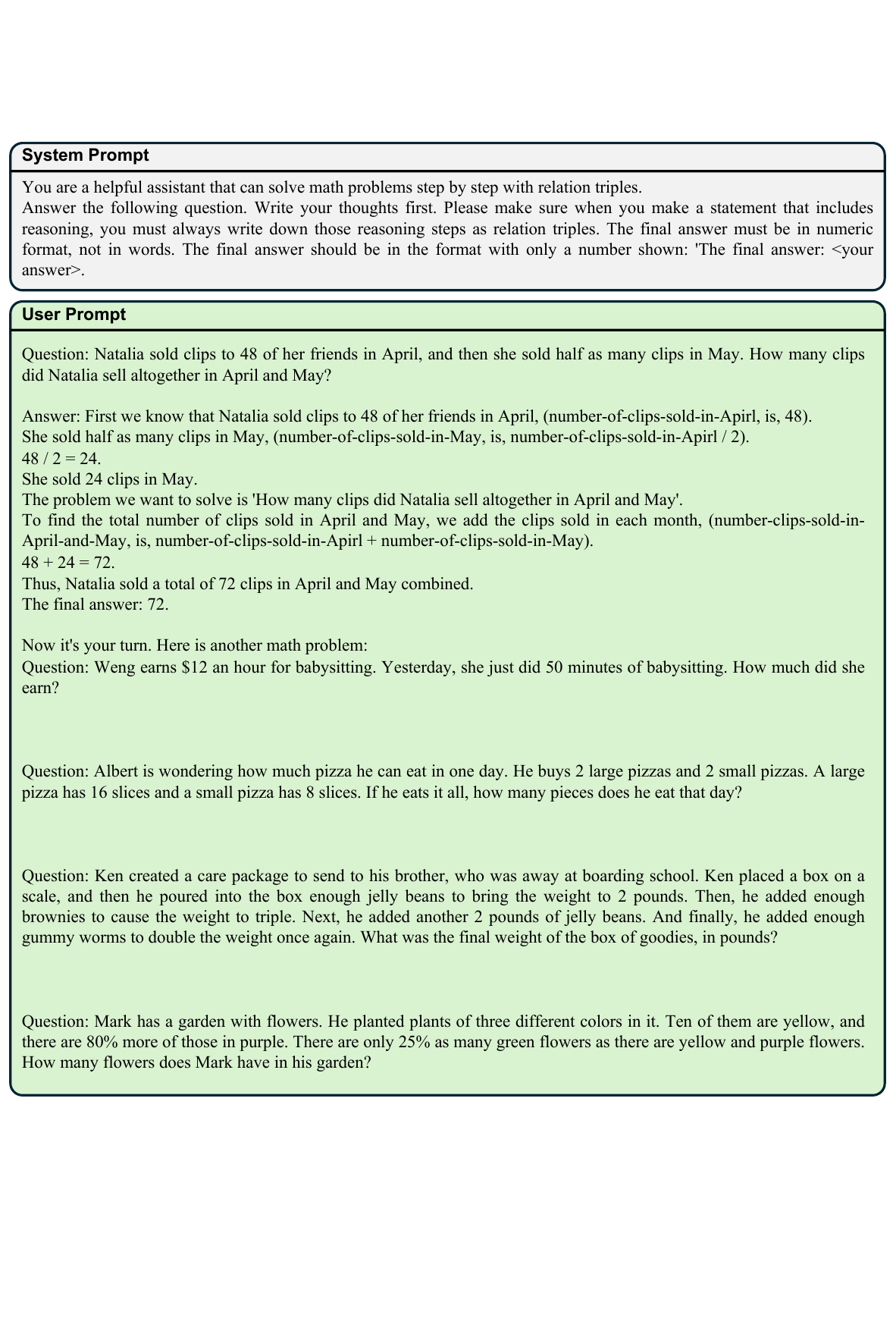}
  \caption {The system and user prompt of the five-shot example in step 1 of our framework.}
  \label{fig:five_shot_step1_prompt_01}
\end{figure*}

\begin{figure*}[h]
  \centering
  \includegraphics[width=1\linewidth]{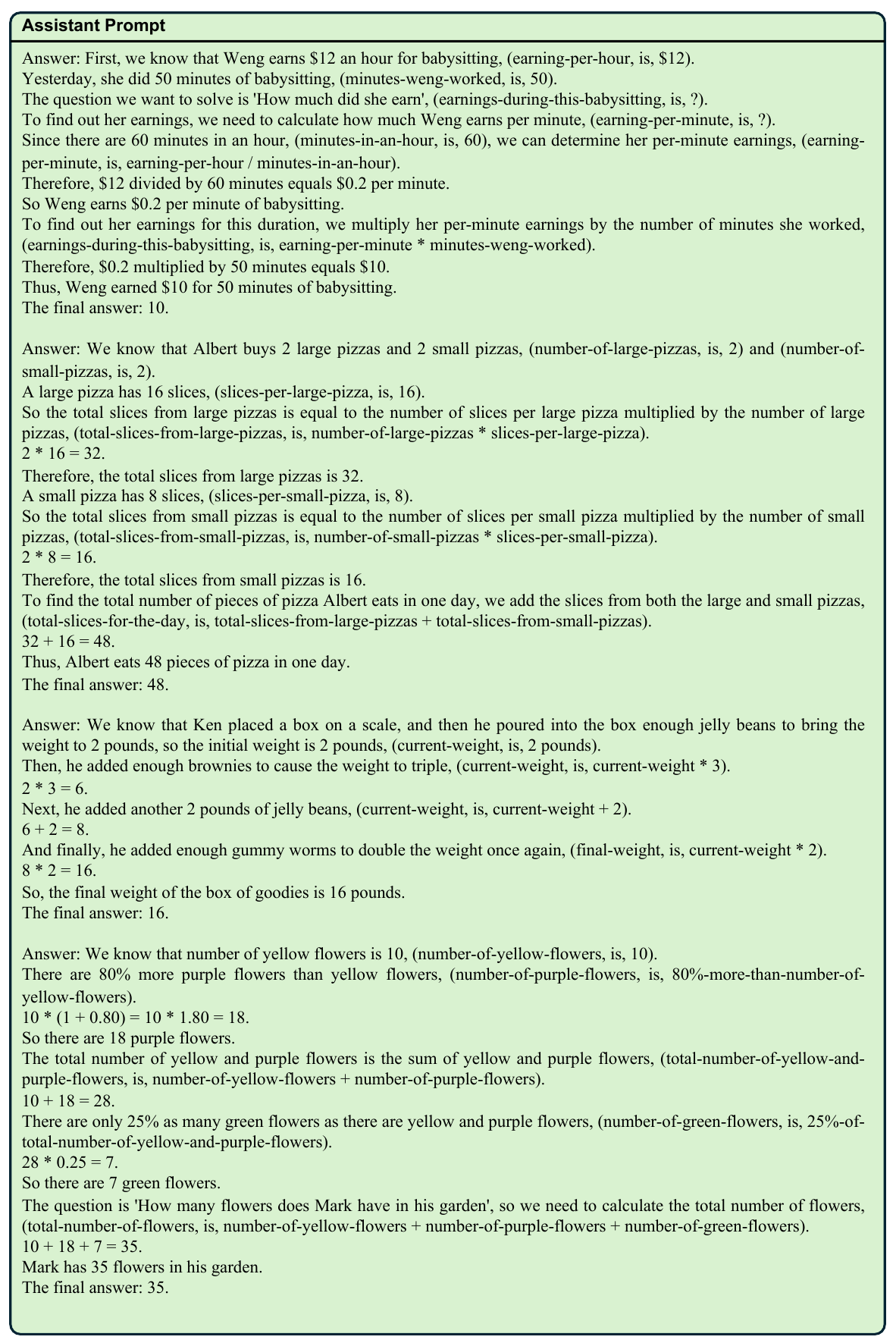}
  \caption {The assistant prompt of the five-shot example in step 1 of our framework.}
  \label{fig:five_shot_step1_prompt_02}
\end{figure*}

Figure~\ref{fig:five_shot_step1_prompt_01} and Figure~\ref{fig:five_shot_step1_prompt_02} show the full prompts used in step 1 of our framework in the five-shot setting.

\begin{figure*}[h]
  \centering
  \includegraphics[width=1\linewidth]{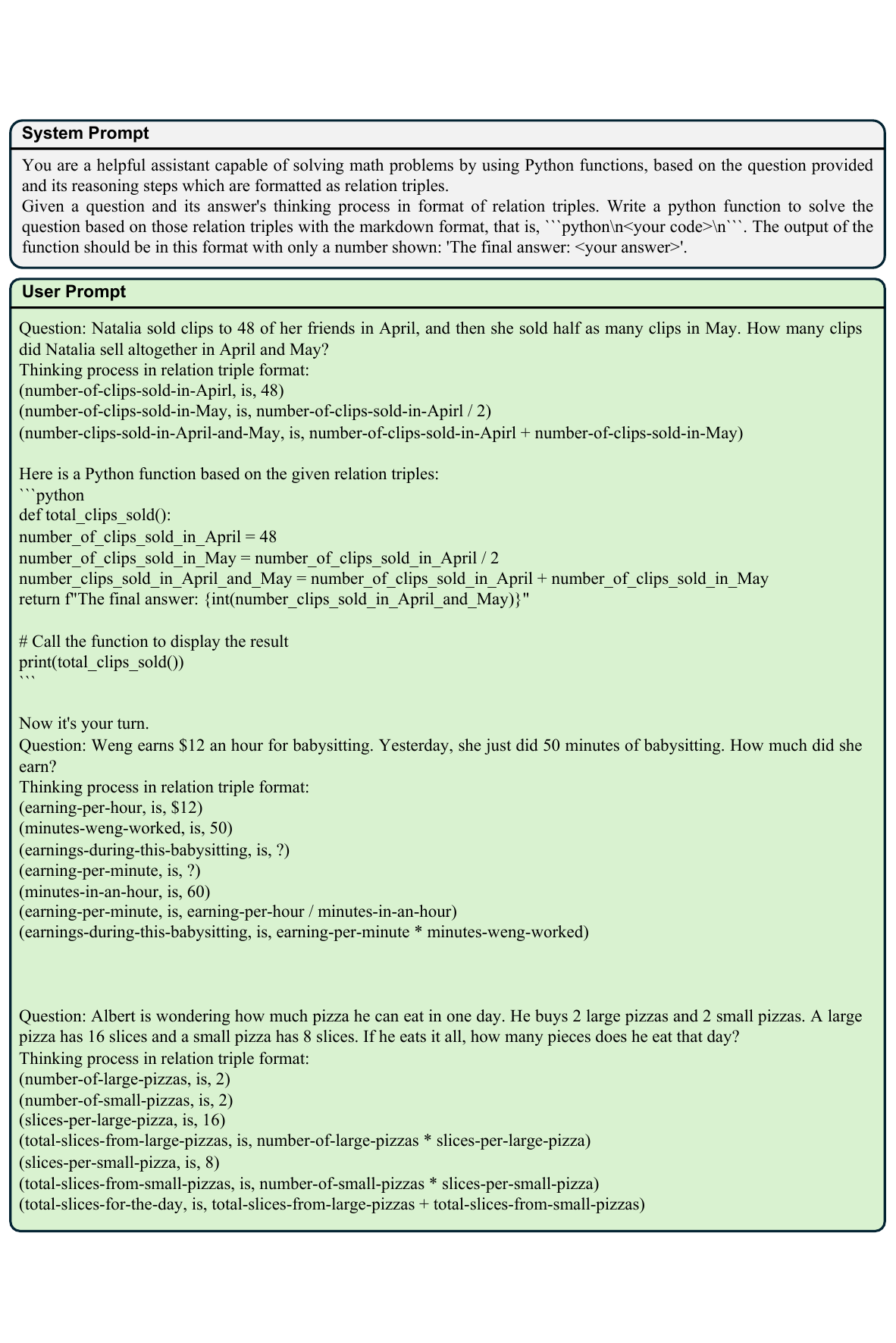}
  \caption {The system and user prompt of the five-shot example in step 2 of our framework.}
  \label{fig:five_shot_step2_prompt_01}
\end{figure*}

\begin{figure*}[h]
  \centering
  \includegraphics[width=1\linewidth]{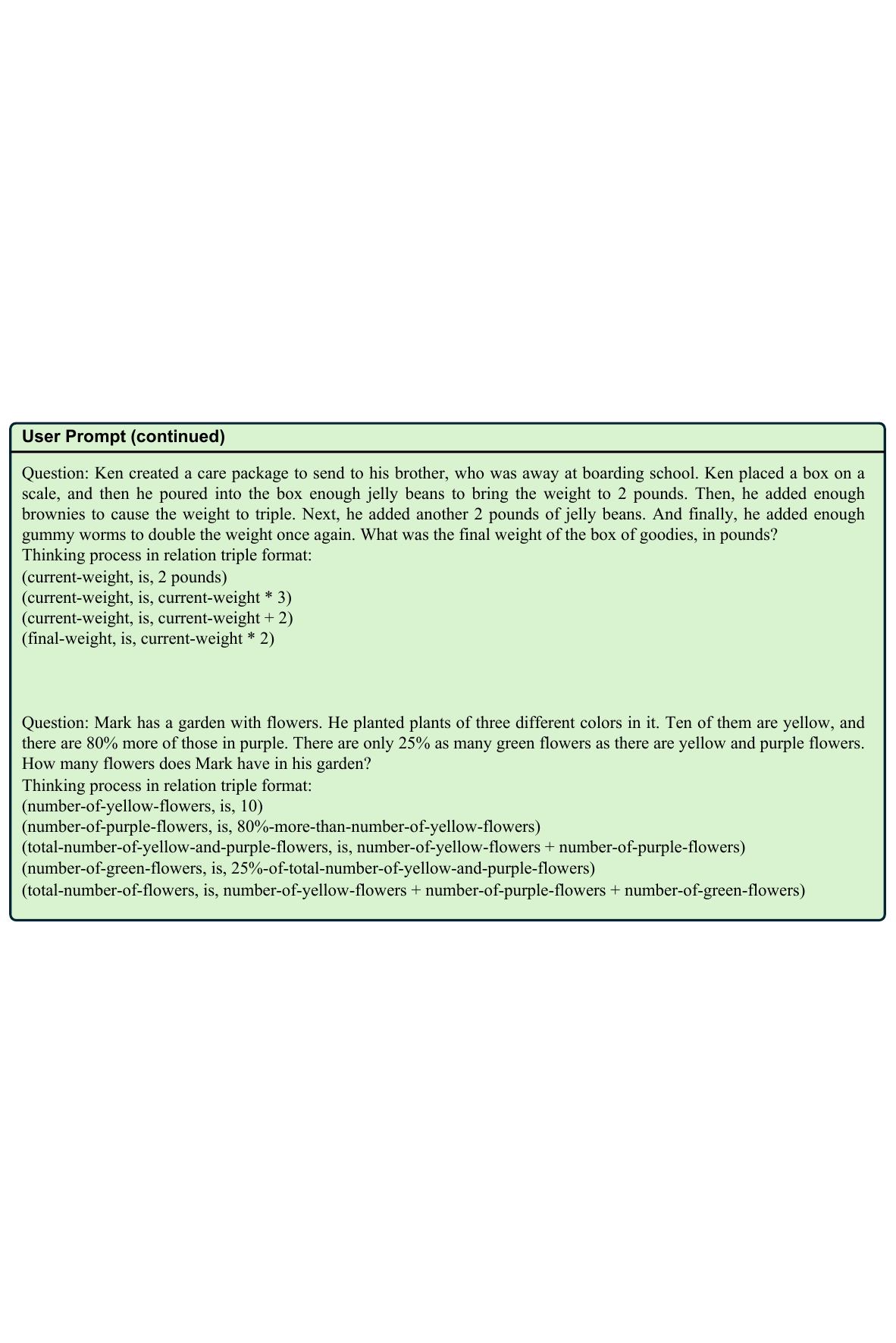}
  \caption {The user prompt of the five-shot example in step 2 of our framework (continued).}
  \label{fig:five_shot_step2_prompt_02}
\end{figure*}

\begin{figure*}[h]
  \centering
  \includegraphics[width=1\linewidth]{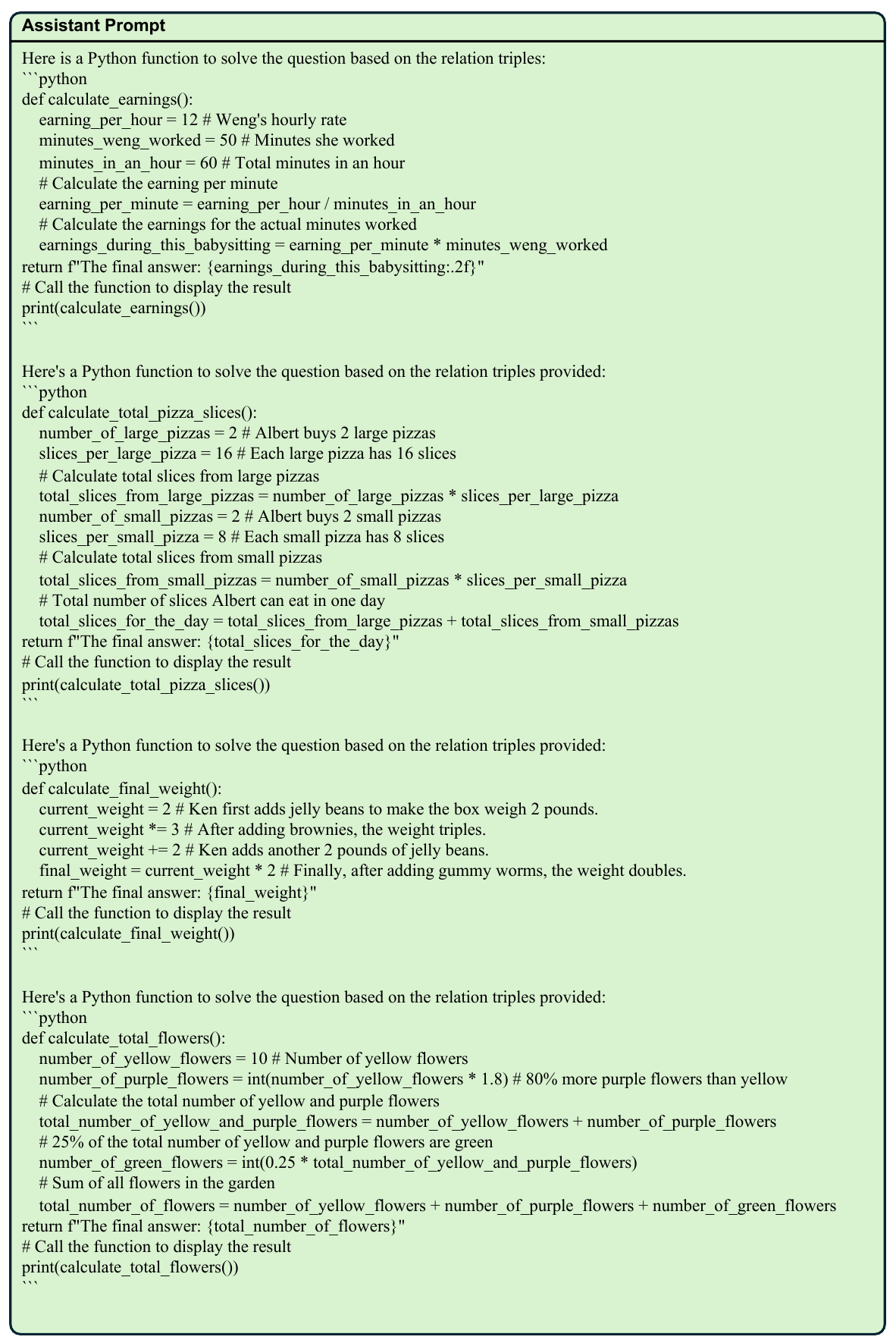}
  \caption {The assistant prompt of the five-shot example in step 2 of our framework.}
  \label{fig:five_shot_step2_prompt_03}
\end{figure*}

Figure~\ref{fig:five_shot_step2_prompt_01}, Figure~\ref{fig:five_shot_step2_prompt_02} and Figure~\ref{fig:five_shot_step2_prompt_03} show the full prompts used in step 2 of our framework in the five-shot setting.

\section{Number of Questions Using Feedback on Arithmetic Datasets}
\label{sec:number_of_questions_using_feedback}
The statistics of the number of the questions which need feedback and the questions which do not need feedback during inference using our framework on GSM8K dataset are shown in Table~\ref{tab:num_questions_feedback_gsm8k}.

The statistics of the number of the questions that need feedback and the questions that do not need feedback in our framework on the other six arithmetic datasets are shown in Table~\ref{tab:num_questions_feedback_other_datasets}.

\begin{table*}[ht]
\centering
\resizebox{\textwidth}{!}{%
\begin{tabular}{@{}cccccc@{}}
\toprule
Backbone & \# No Feedback & \# Feedback (one-time) & \# Feedback (two-times) & \# Feedback (three-times) & \# Feedback \\ \midrule
Llama3-8B-Instruct & 1051 (79.7\%) & 41 (3.1\%) & 20 (1.5\%) & 207 (15.7\%) & 268 (20.3\%) \\
ChatGPT            & 1077 (81.7\%) & 34 (2.6\%) & 43 (3.3\%) & 165 (12.5\%) & 242 (18.3\%) \\
GPT-4o             & 1296 (98.3\%) & 8 (0.6\%)  & 4 (0.3\%)  & 11 (0.8\%)   & 23 (1.7\%)   \\ \bottomrule
\end{tabular}%
}
\caption{Number of questions which require feedback during inference using our framework on the GSM8K dataset.}
\label{tab:num_questions_feedback_gsm8k}
\end{table*}

\begin{table*}[]
\centering
\resizebox{\textwidth}{!}{%
\begin{tabular}{@{}cccccc@{}}
\toprule
Dataset & \# No Feedback & \# Feedback (one-time) & \# Feedback (two-times) & \# Feedback (three-times) & \# Feedback \\ \midrule
SVAMP      & 874 (87.4\%)  & 27 (2.7\%) & 22 (2.2\%) & 77 (7.7\%)   & 126 (12.6\%) \\
ASDIV      & 1746 (83.5\%) & 13 (0.6\%) & 31 (1.5\%) & 302 (14.4\%) & 346 (16.5\%) \\
SingleEQ   & 477 (93.9\%)  & 2 (0.4\%)  & 7 (1.3\%)  & 22 (4.3\%)   & 31 (6.1\%)   \\
SingleOP   & 548 (97.5\%)  & 5 (0.9\%)  & 3 (0.5\%)  & 6 (1.1\%)    & 14 (2.5\%)   \\
AddSub     & 360 (91.1\%)  & 2 (0.5\%)  & 3 (0.8\%)  & 30 (7.6\%)   & 35 (8.9\%)   \\
MultiArith & 590 (98.3\%)  & 4 (0.7\%)  & 4 (0.7\%)  & 2 (0.3\%)    & 10 (1.7\%)   \\ \bottomrule
\end{tabular}%
}
\caption{Number of questions which need feedback during inference using our framework on SVAMP, ASDIV, SingleEQ, SingleOP, AddSub and MultiArith. Note that there are four questions which do not have solutions on ASDIV because program error occurs.}
\label{tab:num_questions_feedback_other_datasets}
\end{table*}




\end{document}